\documentclass[pdflatex,sn-nature]{sn-jnl}% Style for submissions to Nature Portfolio journals
\usepackage{graphicx}%
\usepackage{multirow}%
\usepackage{amsmath,amssymb,amsfonts}%
\usepackage{amsthm}%
\usepackage{mathrsfs}%
\usepackage[title]{appendix}%
\usepackage{xcolor}%
\usepackage{textcomp}%
\usepackage{manyfoot}%
\usepackage{booktabs}%
\usepackage{algorithm}%
\usepackage{algorithmicx}%
\usepackage{algpseudocode}%
\usepackage{listings}%
\usepackage{tabularx}
\usepackage{array}
\theoremstyle{thmstyleone}%
\theoremstyle{thmstyletwo}%

\theoremstyle{thmstylethree}%

\begin{document}

% A title that is up to 15 words in length and free of punctuation, idioms, and puns; the full author list; author affiliation information, including institution, city, and country; and the corresponding author(s) email address
\title[Transferable Self-Harm Surveillance from Emergency Department Triage Notes Using an Evidence-Augmented Machine Learning Approach]{Transferable Self-Harm Surveillance from Emergency Department Triage Notes Using an Evidence-Augmented Machine Learning Approach}


\author*[1]{\fnm{Liuliu} \sur{Chen}}\email{liuliuc@student.unimelb.edu.au}

\author[2]{\fnm{Gowri} \sur{Rajaram}}\email{gowrirajaram96@gmail.com}

\author[2,3]{\fnm{Eleanor} \sur{Bailey}}\email{eleanor.bailey@orygen.org.au}

\author[2,3]{\fnm{Katrina} \sur{Witt}}\email{Katrina.witt@orygen.org.au}

\author[2,3]{\fnm{Michelle} \sur{Lamblin}}\email{Michelle.lamblin@orygen.org.au}

\author[2,3]{\fnm{Jo} \sur{Robinson}}\email{jo.robinson@orygen.org.au}

\author[1]{\fnm{Mike} \sur{Conway}}\email{mike.conway@unimelb.edu.au}

\author[1,4]{\fnm{Vlada} \sur{Rozova}}\email{vlada.rozova@unimelb.edu.au}

\affil*[1]{\orgdiv{School of Computing and Information Systems}, \orgname{University of Melbourne}, \orgaddress{\state{VIC}, \country{Australia}}}

\affil[2]{\orgname{Orygen}, \orgaddress{\state{VIC}, \country{Australia}}}

\affil[3]{\orgdiv{Centre for Youth Mental Health}, \orgname{University of Melbourne}, \orgaddress{\state{VIC}, \country{Australia}}}

\affil[4]{\orgdiv{Centre for Digital Transformation of Health}, \orgname{University of Melbourne}, \orgaddress{\state{VIC}, \country{Australia}}}

% NPJ: No subheadings permitted and up to 150 words in length
% https://www.nature.com/documents/nature-summary-paragraph.pdf

\abstract{
Self-harm is a major public health concern, but current surveillance relying on hospital presentations is inadequate due to the low sensitivity of diagnostic codes. Emergency Department (ED) triage notes, recorded at the initial point of contact, provide a succinct summary of presentations and an opportunity to identify self-harm. We developed a three-stage approach, augmenting traditional machine learning with large language model-based screening and evidence extraction to detect self-harm in ED triage notes. We assessed model transferability across three Australian hospitals. 
Our approach showed AUPRCs of 0.887$\pm$0.016 and 0.884$\pm$0.012 during internal and external validation. Prospectively, it achieved AUPRC of 0.881$\pm$0.008 at the development site, and 0.879$\pm$0.012 and 0.816$\pm$0.015 at two external sites without site-specific retraining. A key advantage of the approach is that it enables identification of the primary self-harm method with an accuracy of 95\%, supporting more granular surveillance beyond binary classification.

% The approach also extracted self-harm methods with 95\% exact-match accuracy
% At the development site, AUPRCs were 0.887$\pm$0.016 on the test set and 0.881$\pm$0.008 on the prospective test set. Without site-specific retraining, AUPRCs were 0.884$\pm$0.012 and 0.816$\pm$0.015 at two external sites, and 0.879$\pm$0.012 on an external-site prospective test set.

% These findings suggest that routinely collected ED triage notes can support scalable and more granular self-harm surveillance across hospital settings.

% Our findings suggest that staged LLM-augmented approaches may support more timely, interpretable, and transferable self-harm surveillance from triage text.
}

\keywords{self-harm surveillance, emergency department, transferability, large language models}



\maketitle

% NPJ: formatting requirement: https://www.nature.com/npjdigitalmed/content-types

\section{Introduction}\label{sec1}

% [SH]
Defined as intentional self-injury (e.g., self-cutting) and/or self-poisoning regardless of motivation or suicidal intent \cite{gillies2018prevalence}, self-harm is a major global public health concern and highly prevalent among young people \cite{hawton2012self}. In Australia, national survey data show 8\% of Australians aged 12-17 years report engaging in self-harm in any 12 months, with higher rates reported by older adolescents aged 16-17 years and by those with mental health disorders \cite{zubrick2016self}. In 2023-2024, hospitalization rates for self-harm were highest among those aged 15–19 (256 per 100,000) \cite{aihw_selfharm_2024}. 
Self-harm is strongly associated with premature mortality, particularly by suicide. An estimated one-in-five of those presenting to emergency departments (EDs) following an episode of non-fatal self-harm will die by suicide within five years \cite{carroll2014hospital}. Self-harm is also frequently repeated, with those engaging in multiple episodes at increased risk of dying by suicide \cite{carroll2014hospital}. 
% Self-harm is also associated with broader family and society impacts \cite{ferrey2016impact}, including increased burden on the healthcare system \cite{kinchin_cost_2020}. 

% [ED-SH: importance for public health surveillance and lit review]
Emergency departments (EDs) are an important setting for identifying self-harm \cite{ross2023emergency}. For many young people, an ED presentation may be their first point of contact with health services during a crisis \cite{gill2017emergency}. ED data therefore have potential value for self-harm surveillance, including tracking changes in presentation volumes over time \cite{silva_characteristics_2024}. 
Capturing specific methods of self-harm in ED surveillance data may be particularly valuable, as this can guide prevention and intervention efforts (including means restriction approaches) and allow for subsequent evaluation of their effectiveness. 
Hospital-based self-harm surveillance systems have been established in several countries, including the United States \cite{kuramoto2017detecting}, the United Kingdom \cite{williams_establishing_2015}, and Australia \cite{hunter, bandara2022surveillance}.

However, self-harm surveillance remains challenging. Existing surveillance systems that rely on diagnostic codes, such as International Classification of Diseases, version 10 (ICD-10), often show low sensitivity (13.8\%-65\%) and may underestimate the occurrence of self-harm and suicide-related admissions \cite{sveticic2020suicidal, randall2017emergency}. This may reflect delayed code assignment, variation in coding practices, and the challenges of translating clinical records into structured diagnostic codes \cite{Malleyicd}.
In contrast, free-text ED triage notes are recorded at the point of care and often contain clinically meaningful descriptions of the presenting problem, making them a potentially valuable source for capturing earlier and more informative surveillance signals.

% [challenge in transferability, and lack of evidence? or only binary?]
Natural language processing (NLP) has created opportunities to use unstructured clinical text for self-harm identification, and has outperformed keyword or code-based approaches \cite{rozova_detection_2021, obeid2020identifying, ayre2021developing}. Obeid et al. reported an AUROC of 0.88 and an F1-score of 0.77 on electronic health records (EHRs) clinical notes \cite{obeid2020identifying}, while Rozova et al. reported an AUPRC of 0.84 using a gradient boosting model on ED triage notes \cite{rozova_detection_2021}.

Despite these advances, model transferability remains relatively underexplored across hospital settings \cite{cusick_portability_2022}. NLP models for suicidality and self-harm that perform well at the development site often show reduced performance when applied to other hospitals \cite{rozova_portability_2025, cusick_portability_2022}. This can require local retraining or adaptation, increasing the need for annotation and implementation resources \cite{barak-corren_validation_2020}. In addition, most existing models focus on binary detection of self-harm behaviours, providing limited insight beyond case identification and little structured supporting evidence \cite{holmes_applications_2025}.

\begin{figure}[t]
    \centering
    \includegraphics[width=\linewidth]{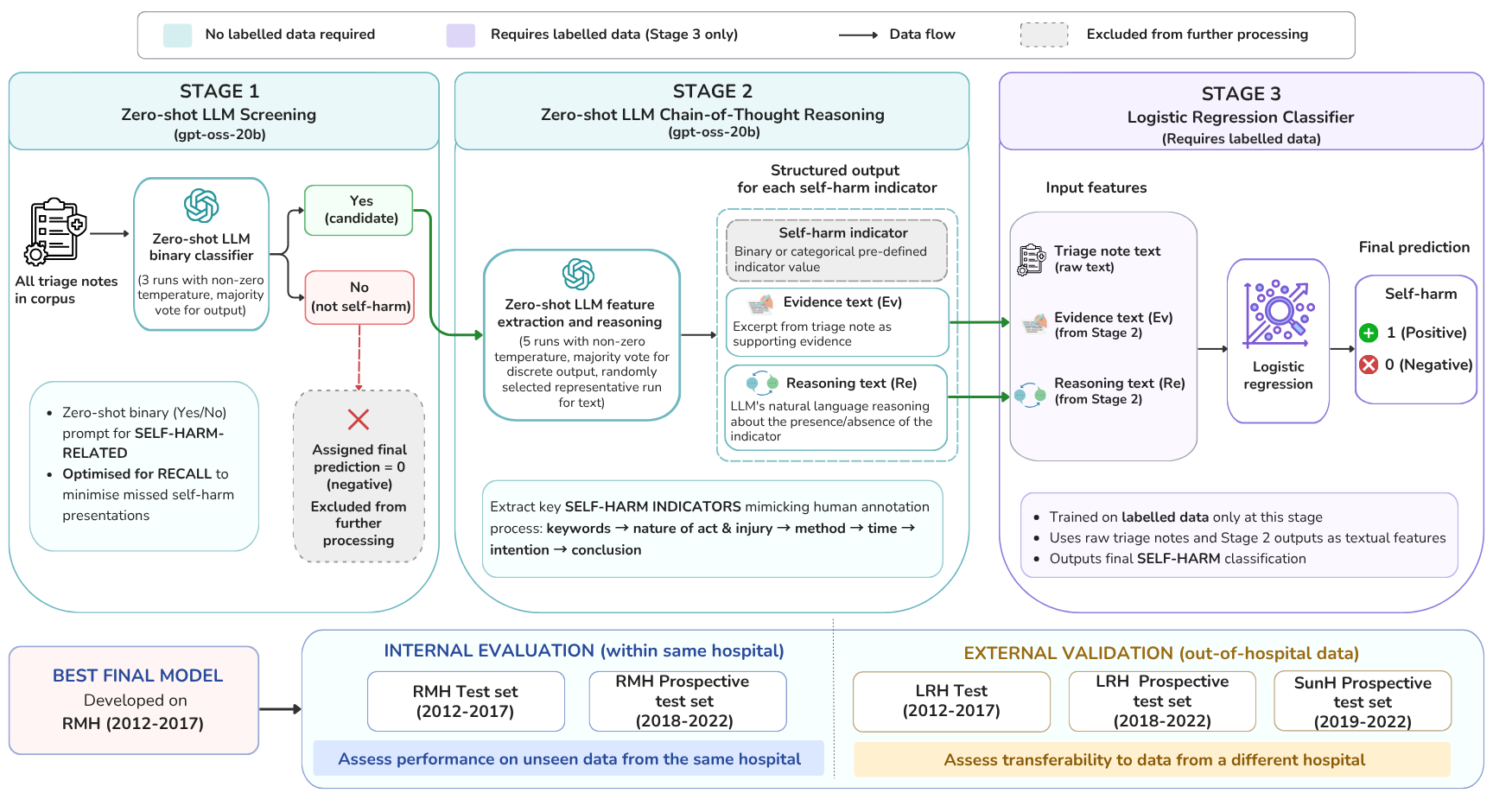}
    \caption{Flowchart of proposed three-stage approach for self-harm surveillance.}
    \label{fig:flowchart}
\end{figure}

% [Our proposed model]
In this study, we propose a three-stage approach combining a large language model (LLM) with a traditional machine learning (ML) classifier to detect self-harm in ED triage notes (Figure \ref{fig:flowchart}). The first stage is designed to screen with high sensitivity for potentially relevant self-harm presentations, significantly reducing the computational burden of subsequent processing. The second stage extracts structured evidence of self-harm from the flagged notes, focusing on the act, injury, method, and intent. The final stage combines the raw note with extracted evidence in an ML classifier to produce the final self-harm prediction. We evaluate binary self-harm detection across both hospital sites and prospective time periods, and separately validate the extracted primary self-harm method against clinician-annotated triage notes.

We developed the approach using 2012–2017 data from the Royal Melbourne Hospital (RMH), a large public tertiary hospital in Melbourne, Victoria, Australia. To assess temporal generalisation, we evaluated the model on later RMH data from 2018–2022. External validation was conducted at two hospital sites: Latrobe Regional Health (LRH), a public regional hospital, and Sunshine Hospital (SunH), a large public hospital in Melbourne’s culturally and linguistically diverse western suburbs \cite{wh_annual_2024}. LRH included both a 2012–2017 external test set and a 2018–2022 prospective test set, allowing assessment of cross-hospital transfer and temporal generalisation. The SunH data span 2019–2022, so its evaluation reflects both cross-hospital and temporal differences relative to the RMH development data. 

% Implication
Overall, this study aims to assess the potential of routinely collected ED triage text for scalable self-harm surveillance and to better understand the challenges of building transferable NLP systems for real-world public health applications.

\section{Results}\label{sec:results}

\subsection{Dataset Descriptions}

Three hospital sites were included in this study (Table~\ref{tab:dataset}). The dataset comprised RMH development, test, and prospective test sets; LRH development, test, and prospective test sets; and SunH development and test sets. 
The RMH development set was divided into four chunks stratified by self-harm label to support sequential pipeline development and prevent data leakage across stages. Records with empty triage text were excluded from all data splits before Stage 1 screening and subsequent analyses, resulting in small differences in the final split sizes shown in Table~\ref{tab:dataset}. External development splits from LRH and SunH were used only for same-hospital upper-bound comparisons.

% SH labels and triage note length
Each note was annotated as self-harm positive or negative by trained suicide and self-harm prevention researchers, with inter-annotator agreement of $\kappa = 0.91$ on a sample of the RMH development data \cite{rozova_detection_2021}. Triage notes were brief across all sites, with median lengths of 21 tokens (IQR 15–27) at RMH, 43 tokens (IQR 32–57) at LRH, and 47 tokens (IQR 37–60) at SunH. The annotated data were highly imbalanced, with self-harm prevalence of 1.36\% at RMH, 1.68\% at LRH, and 1.06\% at SunH.

\begin{table}[t]
\centering
\setlength{\tabcolsep}{3pt}
\caption{Dataset characteristics across hospital sites after exclusion of records with empty triage text. Dev.: development; eval.: evaluation; prev.: prevalence; pos.: positive; prosp.: prospective.}
\label{tab:dataset}
\begin{tabularx}{\linewidth}{>{\raggedright\arraybackslash}l l l l l >{\raggedright\arraybackslash}X}
\hline
\textbf{Site} & \textbf{Period} & \textbf{N} & \textbf{N pos.} & \textbf{Prev.} & \textbf{Role} \\
\hline
RMH   &  &  &  &  &  \\
\quad Dev.  &   2012-2017         & 316{,}895 & 4{,}326 & 1.36\% & Development \\
\qquad Chunk 1 &            & 79{,}200  & 1{,}082 & 1.36\% & Stage 1 prompt dev. \\
\qquad Chunk 2 &            & 79{,}192  & 1{,}081 & 1.37\% & Stage 1 eval.; Stage 2 prompt dev. \\
\qquad Chunk 3 &            & 79{,}268  & 1{,}082 & 1.37\% & Stage 2 eval.; Stage 3 training \\
\qquad Chunk 4 &            & 79{,}235  & 1{,}081 & 1.36\% & Stage 3 model selection/validation \\
\quad Test &    2012-2017           & 79{,}179  & 1{,}082 & 1.37\% & In-distribution test \\
\quad Prosp. test &    2018-2022           & 369{,}148  & 5{,}593 & 1.52\% & prosp. in-distribution test \\
\hline
LRH    &  &  &  &  & External hospital site \\
\quad Dev.    &  2012-2017          & 34{,}234    & 577 & 1.69\% & Stage 3 same-hospital training \\
\quad Test   & 2012-2017         & 136{,}932 & 2{,}306 & 1.68\% & External eval. \\
\quad Prosp. test     &  2018-2022      & 163{,}975 & 2{,}676 & 1.63\% & External prosp. eval. \\
\hline
SunH     &  &  &  &  & External hospital site  \\
\quad Dev.    & 2019-2022             & 60{,}907  & 644 & 1.06\% & Stage 3 same-hospital training \\
\quad Test     &  2019-2022          & 243{,}628 & 2{,}576 & 1.06\% & External eval. \\
\hline
\end{tabularx}
\end{table}

\subsection{Stage-Wise Performance}
The proposed three-stage approach combining LLM-based screening, LLM-based structured evidence extraction, and traditional ML classification is illustrated in Figure \ref{fig:flowchart}. Stage 1 screens all triage notes to identify potential self-harm cases, Stage 2 extracts evidence for five self-harm indicators from the screen-positive notes, and Stage 3 uses the original note and extracted evidence to train a classifier for the final self-harm classification. We first report the performance of each stage.

\subsubsection{Stage 1: Zero-Shot LLM Screening}

Stage 1 applied a zero-shot LLM classifier to all triage notes to identify potential self-harm cases. We prioritised recall to minimise false-negative presentations before progressing to Stage 2. Across all three sites, recall ranged from 0.967 to 0.984 (Table~\ref{tab:stage1}). This screening step reduced the corpus to 4.9\%-6.5\% of all notes, increasing the prevalence from 1.1\%-1.7\% in the full datasets to 21.2\%--26.7\% in the screen-positive sets. This screening process makes subsequent stages computationally more feasible and efficient.

\begin{table}[ht]
\centering
\caption{Stage 1 zero-shot screening: recall and composition of the enriched datasets across sites and time periods. RMH Chunk 2 was used for unbiased prompt evaluation. Prosp.: prospective.}
\label{tab:stage1}
\small
\setlength{\tabcolsep}{2pt}
\begin{tabularx}{\linewidth}{p{3cm} X p{2.3cm} X X X}
\hline
\textbf{Metric} & \textbf{RMH Chunk 2} & \textbf{RMH prosp.} & \textbf{LRH test} & \textbf{LRH prosp.} & \textbf{SunH test} \\
\hline
Recall (95\% CI)
  & $0.974_{\pm 0.010}$
  & $0.967_{\pm 0.005}$
  & $0.984_{\pm 0.005}$
  & $0.983_{\pm 0.005}$
  & $0.976_{\pm 0.006}$ \\
\hline
All triage notes
  & 79{,}192
  & 369{,}148
  & 136{,}932
  & 163{,}975
  & 243{,}628 \\
Screen-positive notes
  & 4{,}699
  & 23{,}350
  & 8{,}849
  & 9{,}871
  & 11{,}835 \\
Reduction to (\%)
  & 5.9\%
  & 6.3\%
  & 6.5\%
  & 6.0\%
  & 4.9\% \\
Post-screen prevalence
  & 22.4\%
  & 23.2\%
  & 25.6\%
  & 26.7\%
  & 21.2\% \\
\hline
\end{tabularx}
\end{table}

\subsubsection{Stage 2: Zero-Shot LLM Structured Evidence Extraction}

\begin{figure}[t]
\centering
\includegraphics[width=.95\linewidth]{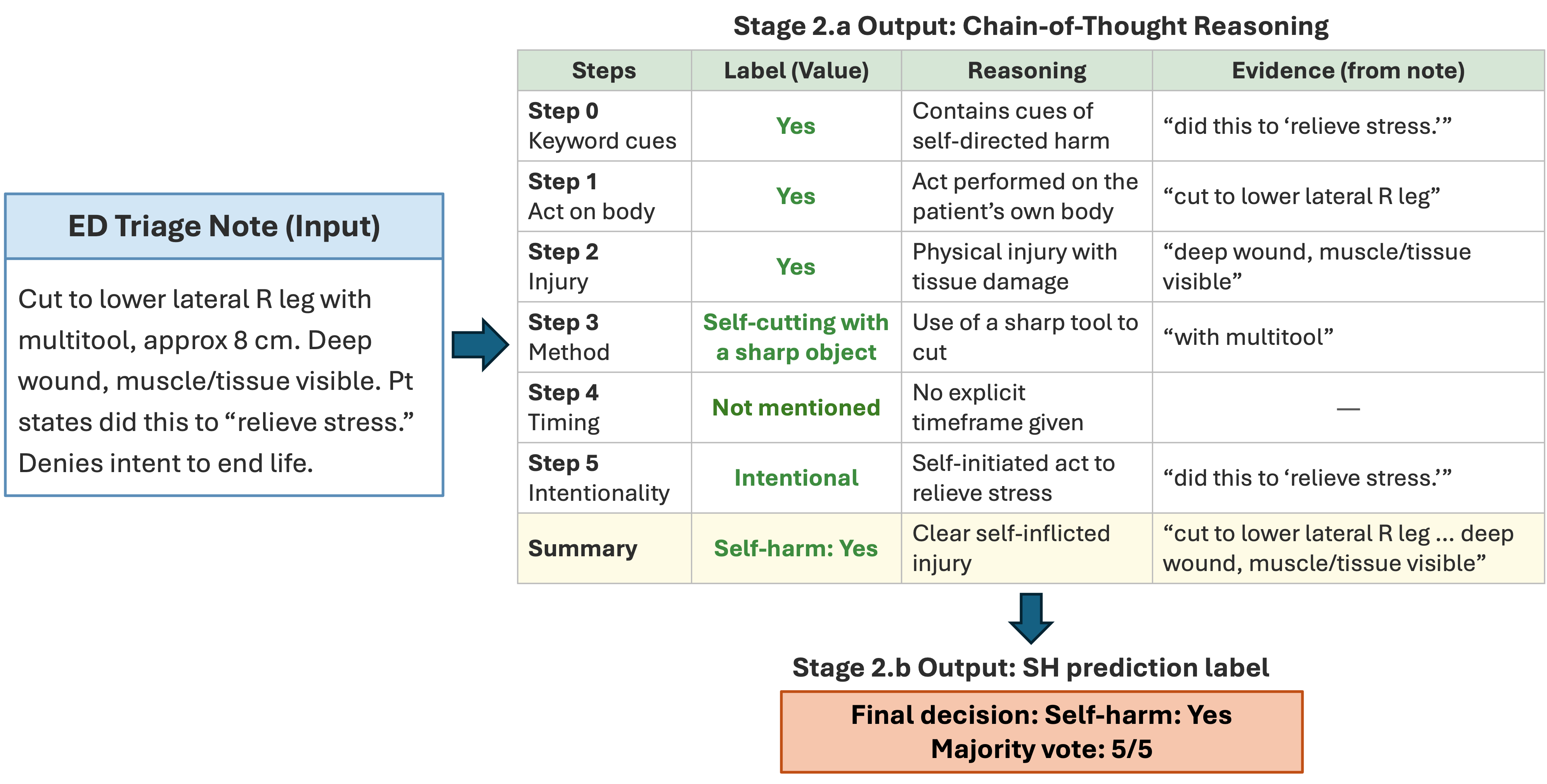}
\caption{Example output of Stage 2. The presented triage note was synthesised for publication purposes. Definitions of each extracted indicator are provided in Methods.}
\label{fig:cot_example}
\end{figure}

Stage 2 applied zero-shot LLM prompting to notes flagged as positive in Stage 1. We instructed the LLM to provide textual evidence and reasoning for five self-harm-related indicators: act, injury, method, timing, and intent, along with self-harm-related keywords and a summary. A binary self-harm label was then generated based on extracted indicators via majority vote across five independent runs. Figure~\ref{fig:cot_example} provides an example of Stage 2 output.

Stage 2 performance on the binary self-harm classification task is summarised in Table~\ref{tab:stage2}. Stage 2  recall across all sites and time periods ranged from 0.933 to 0.961. As a result, when combined together, Stages 1 and 2 maintained high recall ranging from 0.902 to 0.946 across the external and prospective sets. However, Stage 2 demonstrated low precision, substantially lowering the F1 score. This indicates that although the LLM was effective at screening for self-harm presentations, zero-shot binary classification resulted in too many false positives for direct use in surveillance. Instead, we used structured evidence output in Stage 2 as input for a supervised classifier in Stage 3, allowing the final model to calibrate the decision boundary by learning from labelled data.

\begin{table}[ht]
\centering
\caption{Stage 2 performance on the binary self-harm classification task. Performance of Stage 2 alone and in sequential combination with Stage 1. RMH Chunk 3 was used for unbiased prompt evaluation. Prosp.: prospective.}
\label{tab:stage2}
\small
\setlength{\tabcolsep}{4pt}
\begin{tabular}{lccccc}
\hline
& \textbf{RMH Chunk 3} & \textbf{RMH prosp.} & \textbf{LRH test} & \textbf{LRH prosp.} & \textbf{SunH test} \\
\hline
\multicolumn{6}{l}{\textit{Stage 2 alone (applied to screen-positive notes)}} \\
Recall    & $0.933_{\pm 0.015}$ & $0.942_{\pm 0.006}$ & $0.961_{\pm 0.008}$ & $0.932_{\pm 0.009}$ & $0.941_{\pm 0.009}$ \\
Precision & $0.432_{\pm 0.012}$ & $0.452_{\pm 0.006}$ & $0.458_{\pm 0.008}$ & $0.497_{\pm 0.009}$ & $0.389_{\pm 0.007}$ \\
F1        & $0.591_{\pm 0.012}$ & $0.610_{\pm 0.006}$ & $0.620_{\pm 0.008}$ & $0.648_{\pm 0.008}$ & $0.550_{\pm 0.007}$ \\
\hline
\multicolumn{6}{l}{\textit{Stage 1 and Stage 2 combined (applied to full corpora)}} \\
Recall    & $0.902_{\pm 0.019}$ & $0.911_{\pm 0.007}$ & $0.946_{\pm 0.009}$ & $0.917_{\pm 0.011}$ & $0.919_{\pm 0.011}$ \\
Precision & $0.424_{\pm 0.014}$ & $0.442_{\pm 0.007}$ & $0.458_{\pm 0.010}$ & $0.497_{\pm 0.011}$ & $0.389_{\pm 0.008}$ \\
F1        & $0.577_{\pm 0.015}$ & $0.595_{\pm 0.006}$ & $0.617_{\pm 0.009}$ & $0.644_{\pm 0.010}$ & $0.547_{\pm 0.009}$ \\
\hline
\end{tabular}
\end{table}

\subsubsection{Stage 3: Evidence-Augmented Machine Learning Classifier}

Stage 3 was designed to produce the final self-harm predictions. Using textual evidence and reasoning for self-harm indicators from Stage 2, along with the original triage note, we developed a binary supervised classifier. We performed ablation studies to identify the best combination of input features, vectorisers, and classification algorithms. Here, we report the results of feature and model selection. 
\paragraph{Feature Selection}
 
Table~\ref{tab:ablation} demonstrates the performance improvement when augmenting the original triage note with Stage 2 outputs. Adding textual evidence (TF-IDF$_{\text{ev}}$) and reasoning (TF-IDF$_{\text{re}}$) derived in Stage 2 to the original triage notes improved AUPRC from $0.859$ to $0.900$. Compared with textual evidence, LLM reasoning yielded a larger gain in performance (AUPRC of 0.879 vs. 0.897). This suggests that the LLM’s free-text reasoning carried a stronger discriminative signal than the extracted evidence text alone.
 
An exhaustive ablation across all 127 combinations of self-harm indicators found that using all self-harm indicators achieved the highest AUPRC ($0.900$), while \textbf{method + summary} achieved the highest F1 ($0.841$) with a similar AUPRC of $0.896$ (Table~\ref{tab:ablation}). Additional selected results are provided in Supplementary material Table 1. The \textbf{method} indicator captures the primary method of self-harm, while the \textbf{summary} synthesises the model’s overall reasoning, suggesting that these two components provide complementary signals for final classification. Therefore, we selected \textbf{TF-IDF${_\text{note}}$ + MiniLM${_\text{note}}$ + TF-IDF$_{\text{ev+re}}$(method + summary)} as the final Stage 3 feature set.

\begin{table}[t]
\centering
\small
\setlength{\tabcolsep}{2pt}
\caption{Selected feature ablation results on RMH Chunk 4. The classifier was fixed to logistic regression. Additional selected self-harm indicator ablation results are provided in Supplementary material Table 1.}
\label{tab:ablation}
\begin{tabularx}{\linewidth}{X l l l l}
\hline
\textbf{Configuration} & \textbf{AUPRC} & \textbf{F1} & \textbf{Prec.} & \textbf{Rec.} \\
\hline
\multicolumn{5}{l}{\textit{Triage notes alone}} \\
TF-IDF$_{\text{note}}$
  & $0.832_{\pm 0.020}$ & $0.769_{\pm 0.018}$
  & $0.751_{\pm 0.023}$ & $0.788_{\pm 0.024}$ \\
MiniLM$_{\text{note}}$
  & $0.798_{\pm 0.020}$ & $0.713_{\pm 0.022}$
  & $0.757_{\pm 0.025}$ & $0.674_{\pm 0.028}$ \\
TF-IDF$_{\text{note}}$+MiniLM$_{\text{note}}$ (base)
  & $0.859_{\pm 0.017}$ & $0.783_{\pm 0.019}$
  & $0.798_{\pm 0.021}$ & $0.768_{\pm 0.025}$ \\
\hline
\multicolumn{5}{l}{\textit{Triage notes + Stage 2 outputs (pooled across all self-harm indicators)}} \\
Base+TF-IDF$_{\text{ev}}$
  & $0.879_{\pm 0.016}$ & $0.803_{\pm 0.017}$
  & $0.804_{\pm 0.022}$ & $0.802_{\pm 0.024}$ \\
Base+TF-IDF$_{\text{re}}$
  & $0.897_{\pm 0.016}$ & $0.835_{\pm 0.016}$
  & $\textbf{0.852}_{\pm 0.020}$ & $0.818_{\pm 0.023}$ \\
Base+TF-IDF$_{\text{ev}}$ + TF-IDF$_{\text{re}}$
  & $\mathbf{0.900_{\pm 0.016}}$ & $0.837_{\pm 0.015}$
  & $0.831_{\pm 0.020}$ & $\textbf{0.843}_{\pm 0.022}$ \\
\hline
\multicolumn{5}{l}{\textit{Selecting self-harm features for TF-IDF$_{\text{ev+re}}$}} \\
\textbf{Base+TF-IDF$_{\text{ev+re}}$ (method+summary) (final)}
  & $0.896_{\pm 0.017}$ & $\mathbf{0.841_{\pm 0.016}}$
  & $0.849_{\pm 0.020}$ & $0.834_{\pm 0.023}$ \\
\hline
\end{tabularx}
\end{table}

\paragraph{Classifier Comparison}

Table~\ref{tab:classifier} compares the top-performing classifiers using the selected Stage 3 feature set on RMH Chunk 4. The full classifier comparison is provided in Supplementary material Table 2. A simple logistic regression (LR) achieved the best results with the highest AUPRC ($0.896 \pm 0.017$) and F1 ($0.841 \pm 0.016$). In contrast, all three variants of fine-tuned BERT-based models performed slightly worse than the linear models, given the selected feature representation. We selected \textbf{LR} as the final Stage 3 classifier as it provided the best overall performance on the binary self-harm classification task.

\begin{table}[ht]
\centering
\caption{Classifier comparison on RMH Chunk 4. Features fixed as Base+TF-IDF$_{\text{ev+re}}$ (method+summary). All models trained on RMH Chunk 3. Top 3 models shown per classifier family. For the full model comparison, see Supplementary material Table 2.}
\label{tab:classifier}
\begin{tabular}{lllll}
\hline
\textbf{Classifier} & \textbf{AUPRC} & \textbf{F1} &
\textbf{Prec.} & \textbf{Rec.} \\
\hline
\multicolumn{5}{l}{\textit{Traditional machine learning classifiers}} \\
LR
  & $\textbf{0.896}_{\pm 0.017}$ & $\mathbf{0.841_{\pm 0.016}}$
  & $0.849_{\pm 0.020}$ & $0.834_{\pm 0.023}$ \\
MLP
  & $0.886_{\pm 0.016}$ & $0.815_{\pm 0.017}$
  & $0.830_{\pm 0.020}$ & $0.800_{\pm 0.024}$ \\
SVM
  & $0.883_{\pm 0.016}$ & $0.814_{\pm 0.017}$ & $0.793_{\pm 0.020}$ & $\textbf{0.835}_{\pm 0.020}$ \\
\hline
\multicolumn{5}{l}{\textit{BERT-based classifiers (fine-tuned)}} \\
BERT-base
  & $0.871_{\pm 0.017}$ & $0.813_{\pm 0.018}$
  & $\textbf{0.879}_{\pm 0.020}$ & $0.756_{\pm 0.027}$ \\
ClinicalBERT
  & $0.878_{\pm 0.018}$ & $0.817_{\pm 0.017}$
  & $0.826_{\pm 0.021}$ & $0.809_{\pm 0.023}$ \\
SciBERT
  & $0.880_{\pm 0.016}$ & $0.815_{\pm 0.019}$
  & $0.860_{\pm 0.022}$ & $0.774_{\pm 0.026}$ \\
\hline
\end{tabular}
\end{table}

\subsection{End-to-End Performance}
The full three-stage approach was evaluated on in-distribution, external, and prospective test sets. The Stage 3 classifier was trained exclusively on RMH Chunk 3 and applied without site-specific adaptation. We evaluated in-distribution performance on the RMH held-out test set, cross-hospital transfer on the LRH and SunH test sets, and temporal generalisation using prospective RMH and LRH test sets from 2018–2022. SunH records also covered a later period (2019–2022), meaning that SunH evaluation reflects both cross-hospital and temporal differences from the RMH development data.

As shown in Table~\ref{tab:best_model}, the approach achieved strong performance across all three sites. At the development site, AUPRC remained stable from the RMH test set to the RMH prospective test set ($0.887$ vs $0.881$). Cross-hospital performance was also stable at LRH, with AUPRCs of $0.884$ on the LRH test set and $0.879$ on the LRH prospective test set. At SunH, performance was lower, with AUPRC dropping to 0.816 and a lower F1 of 0.772. While transferability was not uniform across hospitals, the recall remained relatively high across all evaluation sets (0.800-0.846).

\begin{figure}[ht]
    \centering
    \includegraphics[width=.9\linewidth]{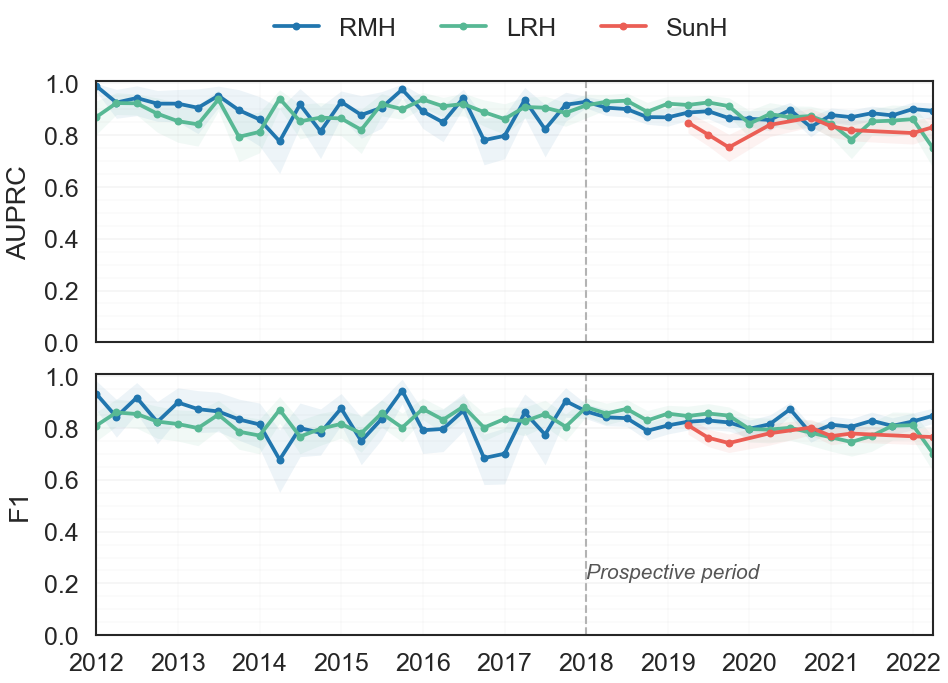}
    \caption{Quarterly AUPRC and F1 across RMH, LRH, and SunH evaluation periods. RMH represents in-distribution evaluation, while LRH and SunH represent cross-hospital validation.}
    \label{fig:quaterly}
\end{figure}

We further examined the temporal changes in performance across the three sites, as shown in Figure~\ref{fig:quaterly}. AUPRC and F1 were calculated over ED presentations grouped by calendar quarter based on the date of arrival. Overall, both metrics remained broadly stable, with no large decline over time. F1 was computed using the fixed classification threshold selected on RMH Chunk 3, and its relatively consistent performance across years and sites suggests that the threshold remained reasonably robust under temporal and geographic shift.

\begin{table}[t]
\centering
\small
\setlength{\tabcolsep}{4pt}
\caption{End-to-end approach performance on held-out and prospective test sets. The Stage 3 classifier was trained on RMH Chunk 3 only, and no site-specific adaptation was applied.}
\label{tab:best_model}
\begin{tabularx}{\linewidth}{>{\raggedright\arraybackslash}l l l l l l}
\hline
\textbf{Metric} & \textbf{RMH test} & \textbf{RMH prosp.} & \textbf{LRH test} & \textbf{LRH prosp.} & \textbf{SunH test} \\
\hline
AUPRC     & $0.887_{\pm 0.016}$ & $0.881_{\pm 0.008}$ & $0.884_{\pm 0.012}$ & $0.879_{\pm 0.012}$ & $0.816_{\pm 0.015}$ \\
AUROC     & $0.985_{\pm 0.005}$ & $0.981_{\pm 0.002}$ & $0.990_{\pm 0.003}$ & $0.989_{\pm 0.003}$ & $0.985_{\pm 0.003}$ \\
F1        & $0.827_{\pm 0.017}$ & $0.823_{\pm 0.008}$ & $0.826_{\pm 0.012}$ & $0.817_{\pm 0.011}$ & $0.772_{\pm 0.011}$ \\
Precision & $0.840_{\pm 0.020}$ & $0.853_{\pm 0.010}$ & $0.813_{\pm 0.014}$ & $0.801_{\pm 0.013}$ & $0.710_{\pm 0.014}$ \\
Recall    & $0.815_{\pm 0.023}$ & $0.800_{\pm 0.011}$ & $0.838_{\pm 0.015}$ & $0.836_{\pm 0.013}$ & $0.846_{\pm 0.014}$ \\
\hline
\end{tabularx}
\end{table}

\begin{figure}
     \centering
     \includegraphics[width=.95\linewidth]{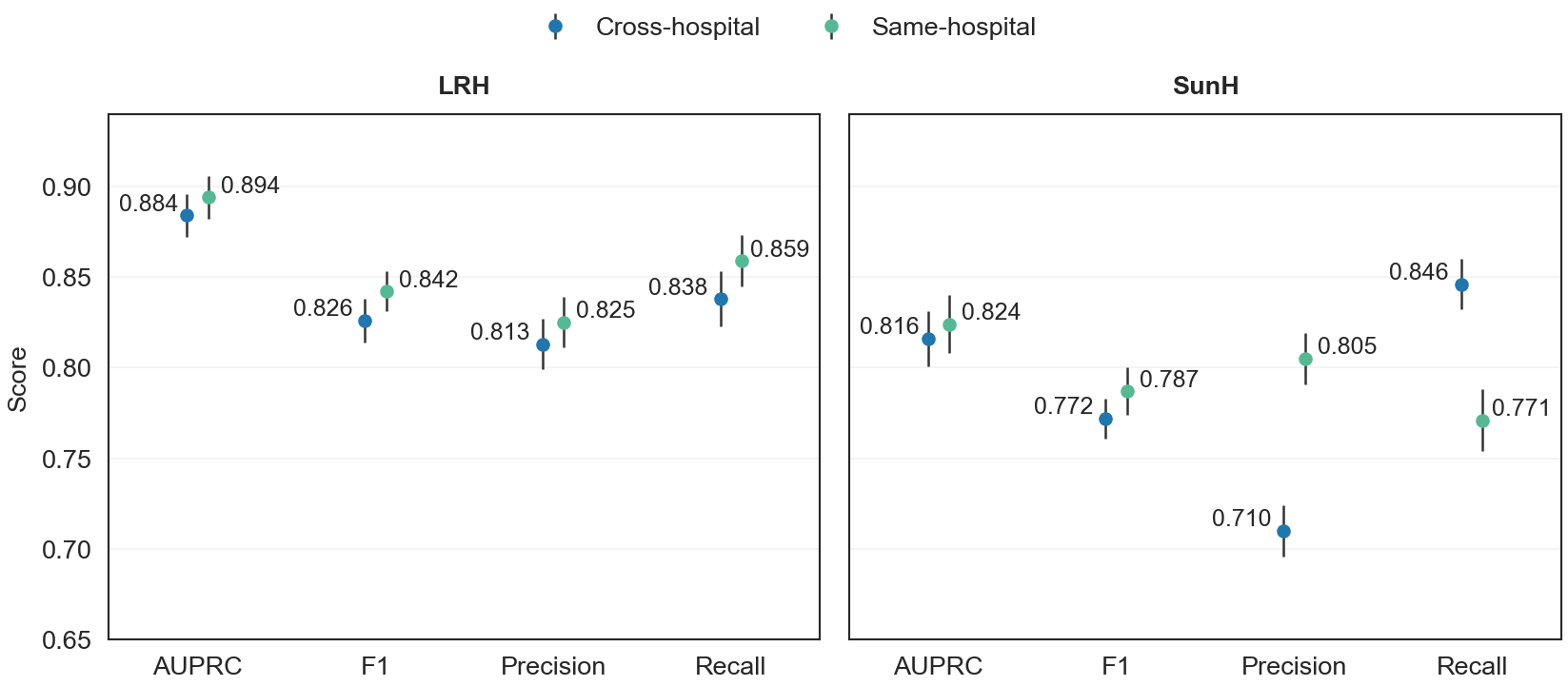}
     \caption{Performance comparison between cross-hospital transfer and same-hospital training. Cross-hospital: Stage 3 classifier trained on RMH Chunk 3, applied without retraining. Same-hospital: Stage 3 classifier trained on local dev set.}
     \label{fig:transfer}
 \end{figure}

\paragraph{Cross-Hospital Transferability}
 We further compare cross-hospital transfer with the same hospital training under our proposed approach, shown in Figure \ref{fig:transfer}. The only difference between settings is the training data for the Stage 3 classifier: in the cross-hospital setting, Stage 3 was trained on RMH Chunk 3, whereas in the same-hospital setting, it was trained on the local development split. At LRH, cross-hospital performance was close to the same-hospital performance, with only small differences. At SunH, cross-hospital performance remained comparable in AUPRC (0.816 vs 0.824) and F1 (0.772 vs 0.787), although precision was lower (0.710 vs 0.805) and recall was higher (0.846 vs 0.771), indicating a different precision-recall trade-off. Overall, these results suggest that the proposed approach retained much of its performance under cross-hospital transfer, particularly at LRH, despite not using site-specific retraining.
 
% \begin{table}[ht]
% \centering
% \caption{Performance comparison between cross-hospital transfer and same-hospital training. Cross-hosp: Stage 3 classifier trained on RMH Chunk 3, applied without retraining. Same-hosp: Stage 3 classifier trained on local dev set.}
% \label{tab:transfer}
% \begin{tabular}{lllll}
% \hline
% \textbf{Metric}
%   & \multicolumn{2}{c}{\textbf{LRH test}}
%   & \multicolumn{2}{c}{\textbf{SunH test}} \\
%   & Cross-hosp & Same-hosp & Cross-hosp & Same-hosp \\
% \hline
% AUPRC
%   & $0.884 \pm 0.012$ & $0.894 \pm 0.012$
%   & $0.816 \pm 0.015$ & $0.824 \pm 0.016$ \\
% AUROC
%   & $0.990 \pm 0.003$ & $0.990 \pm 0.003$
%   & $0.986 \pm 0.004$ & $0.986 \pm 0.004$ \\
% F1
%   & $0.826 \pm 0.011$ & $0.842 \pm 0.011$
%   & $0.772 \pm 0.011$ & $0.787 \pm 0.013$ \\
% Precision
%   & $0.813 \pm 0.014$ & $0.825 \pm 0.014$
%   & $0.710 \pm 0.014$ & $0.805 \pm 0.014$ \\
% Recall
%   & $0.838 \pm 0.015$ & $0.859 \pm 0.014$
%   & $0.846 \pm 0.014$ & $0.771 \pm 0.017$ \\
% \hline
% \end{tabular}
% \end{table}
 
% do we need stage baseline? i.e., directly train using the full corpus?
% existing models? Vlada's model?
\paragraph{Baseline Comparisons}
To assess the contribution of the staged LLM-augmented design, we compared the proposed approach with three baselines trained on the full pre-screening RMH Chunk 3 dataset: (1) logistic regression using TF-IDF and MiniLM features, matching the final classifier without LLM-derived features, (2) ClinicalBERT~\cite{alsentzer2019publicly} fine-tuned with weighted cross-entropy, (3) a gradient boosting baseline (GBM) adapted from the prior ED triage-note self-harm detection approach \cite{rozova_detection_2021}. All baselines used only the raw triage note and did not use LLM-generated structured evidence.

As shown in Table~\ref{tab:baseline_comparison}, the proposed approach achieved the highest AUPRC across all evaluation sets. While performance on RMH was similar to ClinicalBERT, the approach showed clearer advantages on the external sites, particularly at SunH. This suggests that the staged screening and evidence-extraction design may improve robustness under both hospital transfer and temporal shift. Compared with the GBM baseline, the proposed approach achieved higher AUPRC across all held-out and prospective test sets, with particularly larger gains on the external sites.

Beyond classification performance, the proposed approach also provides structured intermediate outputs, including extracted evidence, reasoning, and self-harm method information, which may support more informative surveillance.

\begin{table}[ht]
\centering
\small
\setlength{\tabcolsep}{2pt}
\caption{Baseline comparison: AUPRC across held-out and prospective test sets. All methods were trained on RMH Chunk 3. Note-only baselines used the full pre-screening dataset; the proposed approach used Stage 3 training after Stage 1 screening and Stage 2 evidence extraction.}
\label{tab:baseline_comparison}
\begin{tabularx}{\linewidth}{>{\raggedright\arraybackslash}X l l l l l}
\toprule
Method & RMH test & RMH prosp. & LRH test & LRH prosp. & SunH test \\
\midrule
LR
& $0.816_{\pm 0.021}$
& $0.792_{\pm 0.009}$
& $0.789_{\pm 0.015}$
& $0.783_{\pm 0.014}$
& $0.720_{\pm 0.017}$ \\
GBM (Rozova et al.)
& $0.844_{\pm 0.020}$
& $0.836_{\pm 0.009}$
& $0.772_{\pm 0.017}$
& $0.784_{\pm 0.015}$
& $0.719_{\pm 0.019}$ \\
ClinicalBERT
& $0.880_{\pm 0.019}$
& $0.874_{\pm 0.008}$
& $0.856_{\pm 0.011}$
& $0.856_{\pm 0.012}$
& $0.737_{\pm 0.019}$ \\
\textbf{Proposed approach}
& $\mathbf{0.887_{\pm 0.016}}$
& $\mathbf{0.881_{\pm 0.008}}$
& $\mathbf{0.884_{\pm 0.012}}$
& $\mathbf{0.879_{\pm 0.012}}$
& $\mathbf{0.816_{\pm 0.015}}$ \\
\bottomrule
\end{tabularx}
\end{table}

\paragraph{Computational Cost}
Table~\ref{tab:compute_cost} summarises the computational cost of each stage. Stage 2 required longer inference time per note than Stage 1 as it involved more detailed evidence extraction and multiple inference runs for each screened candidate. However, its overall deployment cost was mitigated by the Stage 1 screening, which reduced downstream processing to approximately 6\% of total notes. At RMH daily presentation volumes of 260--300 notes/day~\cite{rmh_annual_2025}, the full pipeline would require approximately 2.0--2.3 GPU-minutes per day using a single NVIDIA A100 80GB GPU.

\begin{table}[t]
\centering
\small
\setlength{\tabcolsep}{2pt}
\caption{Computational cost of the proposed approach per stage. Energy and CO$_2$ emissions were estimated using CodeCarbon~\cite{benoit_courty_2024_11171501} with Australian national grid emission factors. At RMH daily presentation volumes (260--300 notes/day) \cite{rmh_annual_2025}, the full pipeline requires approximately 2--2.3 GPU-minutes per day.}
\label{tab:compute_cost}
\begin{tabularx}{\linewidth}{>{\raggedright\arraybackslash}X c c c}
\toprule
 & \textbf{Stage 1} & \textbf{Stage 2} & \textbf{Stage 3} \\
\midrule
Model & gpt-oss-20b & gpt-oss-20b & LR \\
Input & All notes & Screen positives ($\sim$6\%) & Screen positives \\
Runs per note & 3 & 5 & 1 \\
Time/note (single run) & 105\,ms & 452\,ms & $<$1\,ms \\
Time/note (all runs) & 315\,ms & 2,260\,ms & $<$1\,ms \\
Hardware & 1$\times$ A100 80GB & 1$\times$ A100 80GB & CPU \\
\midrule
Energy (Wh/1k notes) & 30.2 & 219.3 & negligible \\
Estimated electricity cost (AU\$/1k notes) & 0.009 & 0.066 & negligible \\
CO$_2$ (gCO$_2$e/1k notes) & 16.6 & 120.1 & negligible \\
\bottomrule
\end{tabularx}
\end{table}

\subsection{Clinical Validation of Stage 2 Extracted Self-harm Methods}

Clinical validation was conducted on 1,016 self-harm-positive RMH Chunk 3 notes with valid Stage 2 outputs. A clinical psychologist (EB) independently annotated the primary self-harm method for these notes using a predefined method list (Supplementary material Table 3). To minimise bias, the psychologist reviewed only the original triage notes and was blinded to the LLM-extracted outputs.

\begin{figure}
    \centering
    \includegraphics[width=0.85\linewidth]{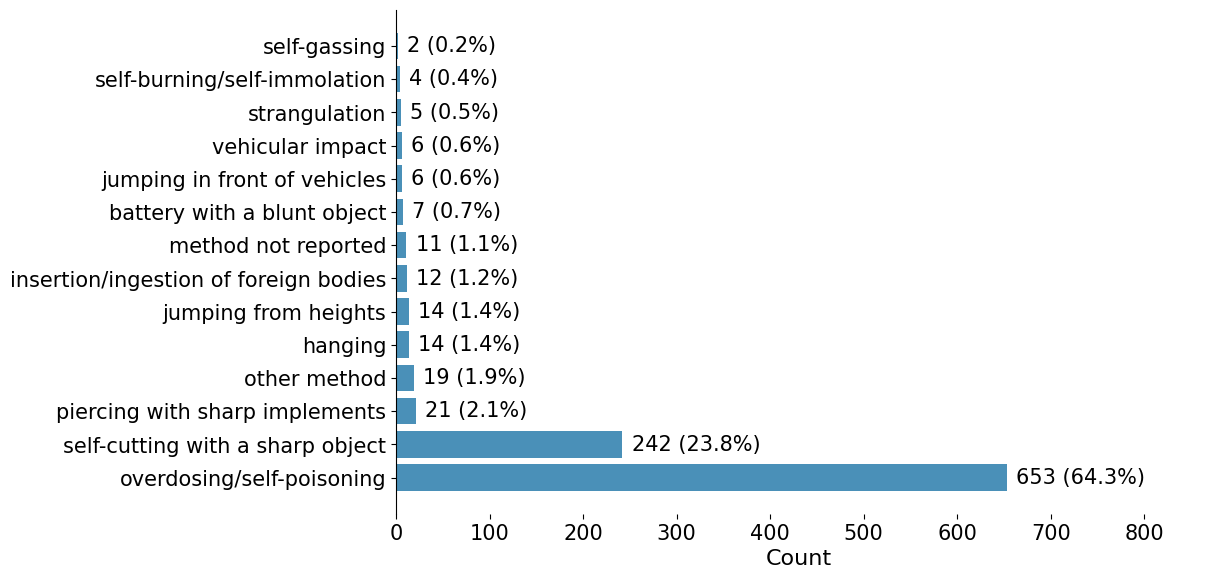}
    \caption{Clinician-annotated distribution of primary self-harm methods in RMH Chunk 3.}
    \label{fig:SH_methods}
\end{figure}

Figure~\ref{fig:SH_methods} shows the clinician-annotated method distribution. Overdosing or self-poisoning was the dominant category (64.3\%), followed by self-cutting with a sharp object (23.8\%); all other method categories were infrequent. 
Across 1016 validated cases, the LLM-extracted method exactly matched the assigned primary method in 965 cases, corresponding to 95.0\% exact-match accuracy. Per-category performance was also strong overall (weighted F1=0.95), although macro F1 was lower (0.76), reflecting reduced performance for rare method categories. Full per-category precision, recall, and F1 scores are reported in Supplementary material Table 4.

Among the 15 predefined method categories, \textit{asphyxia/asphyxiation} was not observed in either the clinician annotations or the LLM-extracted outputs. \textit{Self-burning} (N=4), \textit{overdosing/self-poisoning} (N=653), \textit{hanging} (N=14), and \textit{jumping from heights} (N=14) achieved the highest F1 scores, at 1.00, 0.98, 0.97, and 0.97, respectively. In contrast, \textit{method not reported} (N=11) and \textit{other methods} (N=19) had the lowest F1 scores, at 0.38 and 0.40, respectively, as the LLM often confused these two categories or classified them as \textit{overdosing/self-poisoning}.

% [Disagreement,why]

\begin{figure}[t]
    \centering
    \includegraphics[width=.85\linewidth]{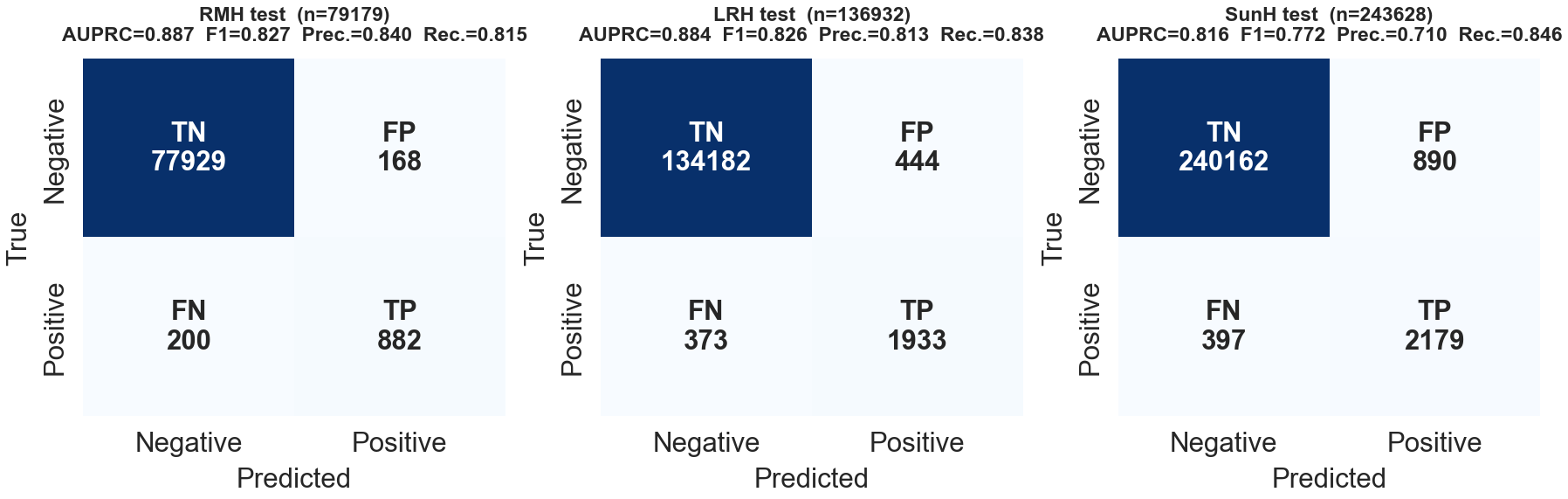}
    \caption{Confusion matrices at the classification threshold of 0.614 for all notes in the test set.}
    \label{fig:cm}
\end{figure}

\subsection{Error Analysis}
To better understand the model’s errors, we examined confusion matrices, predicted probability distributions, and calibration curves across the three held-out evaluation sets: RMH test, LRH test, and SunH test. Because performance on the RMH and LRH prospective test sets remained broadly stable, our error analysis focused on cross-site comparison.

\begin{figure}[t]
    \centering
    \includegraphics[width=.9\linewidth]{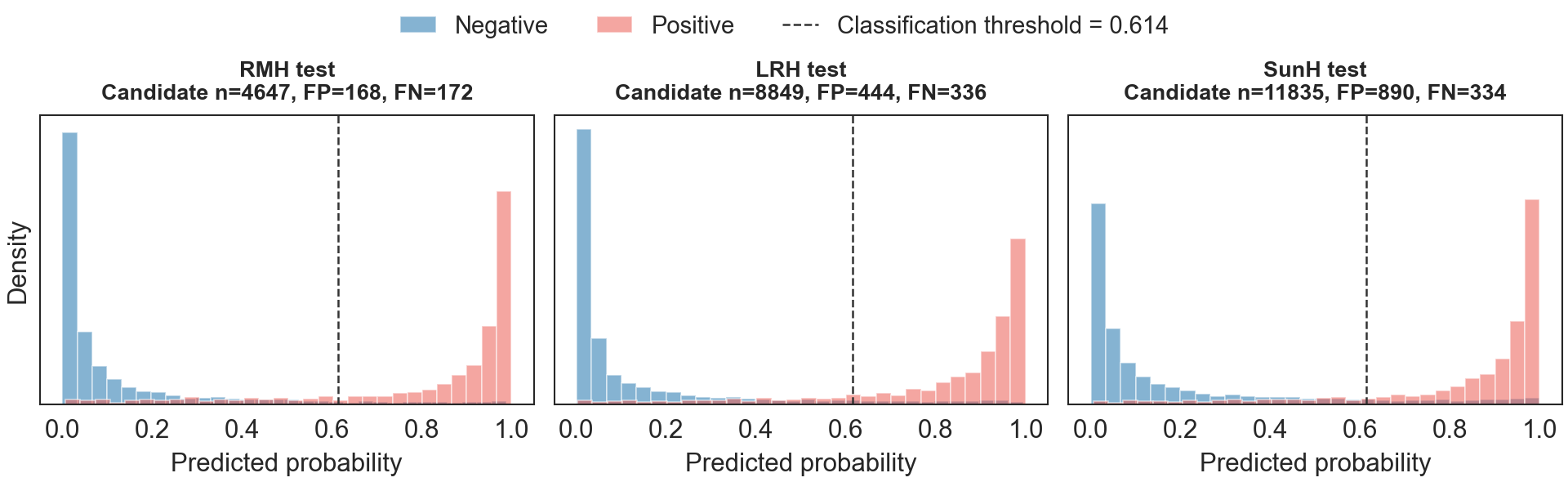}
    \caption{Predicted probability distributions for screen-positive cases at each test site.}
    \label{fig:prob_dist}
\end{figure}

The confusion matrices (Figure~\ref{fig:cm}) showed that most notes were correctly classified at all sites. SunH exhibited a greater burden of false positives than RMH and LRH, consistent with its lower precision. The predicted probability distributions (Figure~\ref{fig:prob_dist}) further suggest that the model generally separated the two classes well, with most positive cases concentrated near 1.0 and most negative cases concentrated near 0. However, the negative class at SunH showed a heavier right tail than at RMH and LRH, with more negative cases receiving moderate predicted probabilities. This pattern is consistent with the lower precision observed at SunH, suggesting that some note patterns at this site activated features associated with self-harm more often than at the other hospitals.

\begin{figure}[t]
    \centering
    \includegraphics[width=.9\linewidth]{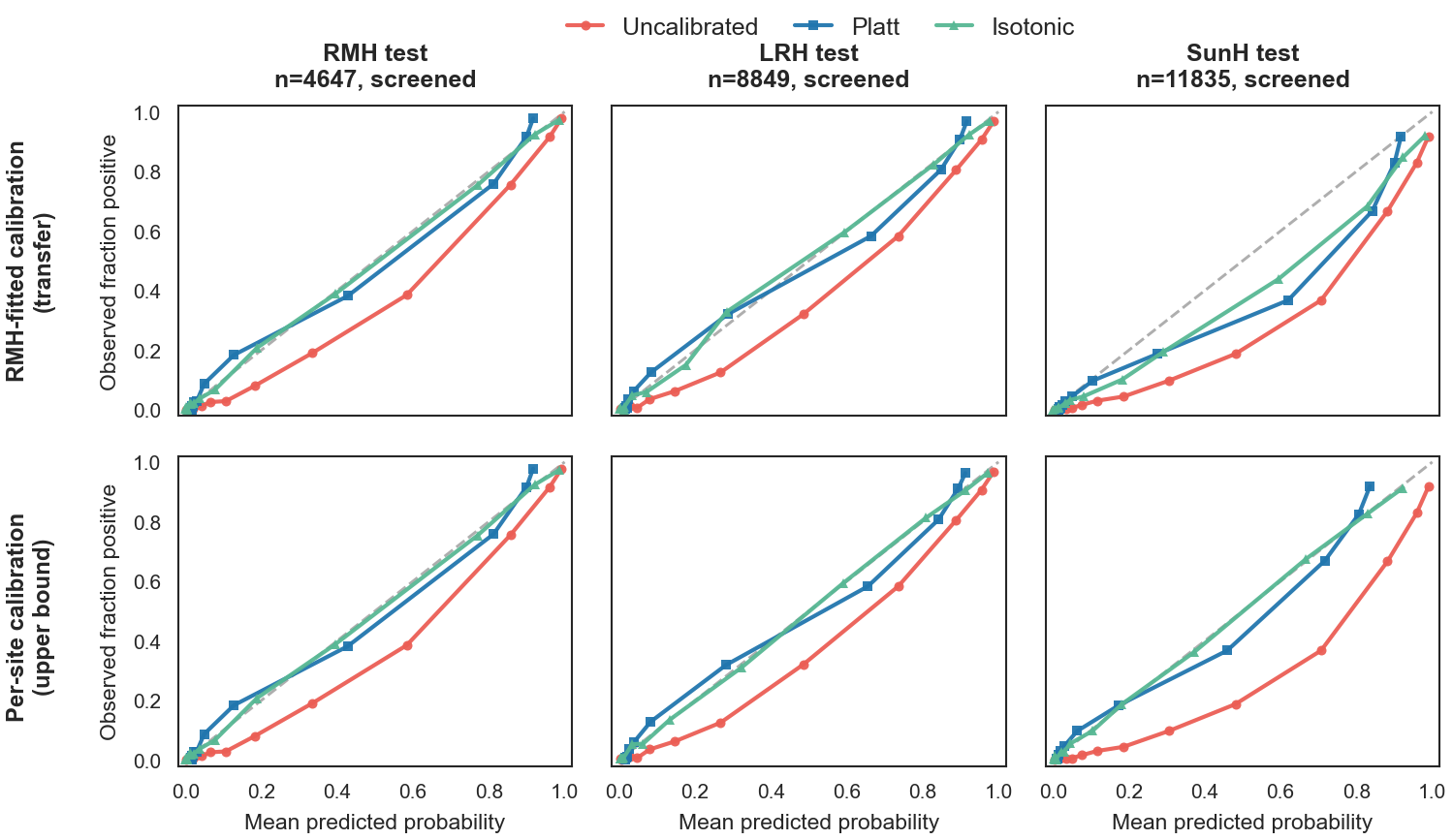}
    \caption{Calibration curves for the final model on screened cases across the held-out test sets. Top row: calibration fitted on RMH and transferred across sites; Bottom row: per-site calibration estimated via 5-fold cross-validation.}
    \label{fig:calibration}
\end{figure}

The calibration curves (Figure~\ref{fig:calibration}) showed that the uncalibrated model did not fully align with the observed positive rates at any site, with the largest mismatch at SunH. Therefore, we compared Platt scaling and isotonic regression fitted on RMH, simulating transfer without local labels, and also examined per-site calibration as an upper bound. RMH-fitted Platt scaling transferred reasonably to LRH but less well to SunH. Per-site isotonic calibration improved alignment between predicted probabilities and observed positive rates at all sites. Although calibration was not incorporated into the final model, these findings suggest that site-specific recalibration may improve probability reliability and may help achieve a more suitable precision–recall balance in future development.

% [qualitative review]
To gain qualitative insights regarding the limitations of our classifier, we reviewed the 30 highest-confidence and 30 near-threshold false positives at each site. The review was conducted by the first author (LC) using the study annotation rules and was used only to describe error patterns. 

Table \ref{tab:fp_examples} presents synthetic examples illustrating the false-positive error patterns. High-confidence false positives fell into two broad categories. The first involved cases where the reference label itself may be debatable. For example, some SunH false positives described overdose with documented suicidal intent (Table \ref{tab:fp_examples}, No.1), which the model classified as self-harm but were labelled negative in the reference data. This suggests possible reference-label disagreement. The second category involved cases that contained self-harm cues but were labelled negative under the annotation scheme. These included self-harm described as part of a previous rather than the current ED presentation (No.2), suicidal ideation or threats without a self-harm act (No.3), and overdose cases assigned to a mutually exclusive alcohol/other-drug overdose category when the Glasgow Coma Scale (GCS) score met predefined criteria (GCS$<$9) (No.4). In these cases, the model responded to genuine self-harm language in the note, but the annotation rules excluded them for specific reasons. Near-threshold false positives were more often clearer negatives, including recreational drug use (No.5) or insufficient evidence of injury (No.6).

\begin{table}[t]
\centering
\small
\setlength{\tabcolsep}{2pt}
\caption{Synthetic examples of false-positive patterns observed during qualitative review. Examples are synthetic and do not represent real patient records. Each example illustrates a recurring pattern observed across reviewed cases.}
\label{tab:fp_examples}
\begin{tabularx}{\linewidth}{>{\raggedright\arraybackslash}p{0.7cm} >{\raggedright\arraybackslash}p{1.8cm} >{\raggedright\arraybackslash}X >{\raggedright\arraybackslash}p{3.8cm}}
\toprule
\textbf{No.} & \textbf{Category} & \textbf{Synthetic triage note} & \textbf{Pattern} \\
\midrule
1 & High-confidence 
& Patient felt anxious at triage, 2000hours took intentional overdose of 6 x 500mg paracetamol + 7 x 30mg diclofenac tablets with suicide intent, social/family stressors. 
& Possible reference-label disagreement for overdose with suicidal intent 
\\
2 & High-confidence 
& Represents after being discharged this am post self-inflicted lacerations to l) wrist. now states of feeling cold and nausea. 
& Self-harm not linked to current ED presentation 
\\
3 & High-confidence 
& Held knife to the neck stating killing self. Girlfriend called police. 
& Suicidal ideation or threat without self-harm action 
\\
4 & Near-threshold 
& polypharmacy od - ice, heroin, \& etoh. agitated, gcs 8.
& Alcohol/other-drug overdose category based on GCS
\\
5 & Near-threshold 
& Taken mdma \& cannabis, found unconscious. Now GCS 13. phx. ivdu. 
& Plausible recreational drug use
\\
6 & Near-threshold 
& Suicidal attempt by hanging. Nil marks/bruising noted, c/o sore throat, normal voice.
& Insufficient evidence of injury 
\\
\bottomrule
\end{tabularx}
\end{table}

% No subheadings, limitations, or conclusions sections are permitted
\section{Discussion}\label{sec12}

% main finding: framework, cross-hospital transferability
This study investigates two key questions in self-harm surveillance: whether ED triage text can support both accurate binary classification and structured evidence extraction for self-harm presentations, and whether such an approach can maintain performance across hospital sites and time periods. To address these, we developed a three-stage approach combining high-sensitivity LLM screening, structured LLM-based evidence extraction, and a final evidence-augmented classifier. 
The final model trained on RMH Chunk 3 achieved AUPRCs ranging from 0.816 to 0.887 across all evaluation sets (Table \ref{tab:best_model}), with stable performance on prospective test sets at both the development site (RMH) and LRH, suggesting robustness to temporal shift. Cross-hospital transfer was strong at LRH, where performance was close to the same-hospital upper bound, but weaker at SunH. At SunH, although cross-hospital and same-hospital AUPRCs were similar (0.816 vs 0.824), precision was notably lower under cross-hospital transfer (0.710 vs 0.805).

% justify the stage design is effective overall
The proposed approach effectively separates high-sensitivity screening, structured evidence extraction, and final classification. Stage 1 reduced the full corpus to a smaller high-recall candidate set, which reduces class imbalance and the computational burden of downstream processing. Stage 2 transformed brief triage text into explicit self-harm indicators with supporting evidence and explanations. Feature ablation showed that adding these LLM-derived outputs improved performance over the note-only base representation, indicating that extracted evidence and reasoning provided useful information beyond the raw note itself. This may be particularly valuable for ED triage text, where documentation is often brief, inconsistent, and highly abbreviated. Additionally, since Stage 2 outputs were generated using a fixed prompt and model, they may provide a more standardised representation of self-harm cues than raw triage notes alone, potentially reducing the final classifier’s reliance on site-specific wording.

% justify hte extracted SH features, esp SH methods, the ft combination for clf (method + summary) in stage 3
The extracted self-harm indicators are also an important part of the proposed approach. In consultation with self-harm and suicide prevention researchers, we defined five indicators (act, injury, method, intent, and timing) with a final summary step to reflect key elements of the annotation process. For each indicator, the approach generated structured intermediate outputs, including extracted evidence, reasoning, and feature labels. The \textit{method + summary} feature combination achieved the highest F1 (0.841) with an AUPRC of 0.896, compared with the highest AUPRC of 0.900 when using all indicators. Method information is concrete and often highly discriminative, while the summary captures broader contextual interpretation. Clinical validation further showed that the LLM-extracted self-harm method had high agreement with clinician annotation, achieving 95.0\% accuracy. These findings suggest that structured LLM outputs may support both final classification and more granular surveillance of self-harm presentations.
However, only the self-harm method labels were clinically reviewed in this study, and further validation is needed for the evidence and reasoning outputs and for the remaining extracted indicators.

% justify simple model works than bert and llm
Interestingly, the classifier comparison showed that a simple logistic regression model with evidence-augmented features outperformed both fine-tuned BERT-based models in Stage 3 and the direct LLM-generated prediction from Stage 2. One possible explanation is that the selected feature space already contained explicit evidence and reasoning signals extracted by the LLM, reducing the need for a more complex contextual model. In this setting, a simpler linear classifier may have been sufficient to use these features effectively and may have been less sensitive to site-specific wording patterns. 
Although the approach still relies on LLM inference at Stages 1 and 2, using a lightweight final classifier makes the output more interpretable and easier to inspect, retrain, and adapt to local settings. In practice, this may allow health services to retain a fixed LLM-based extraction process while adjusting the final classifier or threshold to adapt to local documentation practices and surveillance priorities.

% SunH precision and prevalence calibration
However, the SunH result requires particular attention. Compared to RMH and LRH, the AUPRC was lower at SunH overall, but the same-hospital and cross-hospital AUPRCs were similar (0.824 vs 0.816). This indicates that the lower performance was not solely due to the training on RMH data. The main difference under the cross-hospital transfer was a shift in precision–recall balance, with lower precision (0.710 vs 0.805) and higher recall (0.846 vs 0.771). This may reflect institutional differences in documentation style, patient demographics, and local triage practice. 
The qualitative false-positive review suggested that some SunH false positives were overdose presentations with documented suicidal intent, indicating possible reference-label disagreement in a subset of cases.
SunH also covered a later period than the RMH development set, although the stable RMH and LRH prospective results suggest that the lower SunH performance is not likely to be explained by temporal shift alone. 
From a surveillance perspective, this may still be acceptable if sensitivity is prioritised, but it also indicates that threshold recalibration may help tailor performance to local needs.

%Surveillance application and implications
Our proposed approach demonstrates the potential of using ED triage notes for self-harm surveillance, reducing the reliance on diagnostic codes while avoiding fully black-box classification.
By classifying directly from ED triage notes, the approach supports more accurate identification of self-harm presentations and improved case ascertainment. Extracted structured intermediate outputs, including evidence and reasoning of self-harm indicators, further support audit, review of uncertain cases, and surveillance of method-specific patterns. The cross-hospital findings suggest that this approach can support surveillance across sites without requiring extensive local annotation and retraining, although threshold recalibration may still be needed in some settings.

%Limitations and future work
Several limitations should be considered. First, the study was conducted on three hospital sites within one Australian state, and broader evaluation across other health systems will be needed to further assess transferability. Second, the approach was developed and evaluated using only a single LLM (\texttt{gpt-oss-20b}), and performance may vary with different LLMs. Future work should evaluate whether smaller or more efficient LLMs can provide comparable performance. Third, although the approach provides interpretable self-harm features, only the self-harm method labels were clinically validated. Fourth, although the approach substantially reduced downstream processing, LLM-based inference still requires GPU resources and has associated energy and carbon costs. Finally, this was a retrospective study of surveillance potential rather than a deployment study, and further evaluation would be needed before use in surveillance practice.

% Conclusion
In summary, this study presents a three-stage approach combining LLM screening, structured evidence extraction, and evidence-augmented classification for self-harm surveillance from ED triage notes. The approach captured key indicators of self-harm presentations and demonstrated strong performance under temporal and cross-hospital evaluation. Our findings suggest that ED triage text can support scalable and more granular self-harm surveillance across hospital settings and over time.

\section{Methods}\label{sec:methods}

\subsection{Study Design and Ethical Approval}
% Ethical approval
%The Self-Harm Monitoring System for Victoria received ethical approval from the Melbourne Health Human Research Ethics Committee (HREC; 2017.342).

This study used a retrospective analysis of routinely collected emergency department triage records from three Australian hospital sites. Ethical approval was granted by the Melbourne Health Human Research Ethics Committee (HREC; 2017.342). Patient data were de-identified prior to analysis and no individual-level identifiers were used.

\subsection{Data and Annotation}

\subsubsection{Data Sources}

Triage notes were extracted from patient management systems at three hospital sites in Victoria, Australia: the Royal Melbourne Hospital (RMH), Latrobe Regional Health (LRH), and Sunshine Hospital (SunH). RMH and LRH records span 2012–2022, and SunH records span 2019–2022. Each record comprises a brief free-text triage note entered by nursing staff at the point of care, alongside demographic and administrative fields. Only the free-text triage note was used as model input.

RMH is one of Australia's leading public tertiary hospitals~\cite{rmh_annual_2025}. LRH is the principal referral hospital for the Gippsland region, serving a predominantly rural and regional catchment~\cite{lrh_annual_2025}. SunH is a public hospital in Melbourne’s western suburbs and part of Western Health, a health service serving one of Australia’s fastest-growing and most diverse regions~\cite{wh_annual_2024}.

% Look into the PLOS paper
\subsubsection{Self-Harm Annotation}
Ground truth self-harm labels were coded independently by pairs of trained research assistants experienced in suicide and self-harm prevention research, with consensus discussions involving a third member of the research team (KW) to resolve unclear cases. Inter-annotator agreement on a sample of the RMH development data was $\kappa = 0.91$, as reported in \cite{rozova_detection_2021}. A presentation was labelled as self-harm positive if the triage note indicated intentional self-injury or self-poisoning, regardless of motivation or suicidal intent.

\subsection{Proposed Three-Stage Approach}

The proposed approach consisted of three sequential stages (Figure~\ref{fig:flowchart}). Stage 1 used a zero-shot LLM to screen all triage notes for potential self-harm presentations. Stage 2 applied zero-shot chain-of-thought prompting to the screen-positive notes to extract interpretable evidence for self-harm indicators. Stage 3 combined the original triage note with the Stage 2 outputs as inputs to a supervised classifier to generate the final prediction. Only Stage 3 required labelled training data.

\paragraph{Stage 1: Zero-Shot LLM Screening}
Stage 1 is a high-sensitivity screening stage designed to identify potentially self-harm-related presentations from the full triage corpus. Its purpose is to retain ambiguous but plausible self-harm cases for further analysis, while substantially reducing the number of notes requiring downstream processing. 
The model returned one of three labels for self-harm related: \texttt{yes}, \texttt{no}, or \texttt{unsure}. To minimise missed self-harm presentations, notes labelled \texttt{yes} or \texttt{unsure} were both retained for Stage 2. Notes labelled \texttt{no} were excluded from further processing and assigned a final prediction of self-harm negative. 

\paragraph{Stage 2: LLM-Based Structured Evidence Extraction}
Stage 2 is a structured reasoning stage applied to the notes retained after Stage 1 screening. Rather than producing only a binary judgement, it extracts evidence and reasoning for key self-harm indicators from each note. This provides a more granular and interpretable representation of self-harm for downstream classification. In consultation with self-harm and suicide prevention researchers, we defined five key self-harm indicators with a final summary step, as shown in Table~\ref{tab:stage2_steps}. 
For each indicator, the model produced three outputs: supporting evidence text extracted from the triage note, free-text reasoning, and a feature label (binary or categorical).

\begin{table}[t]
\centering
\caption{Self-harm indicators and reasoning steps extracted in Stage 2. For each indicator, the LLM produced supporting evidence, free-text reasoning, and a feature label.}
\label{tab:stage2_steps}
\begin{tabularx}{\linewidth}{llX}
\hline
\textbf{Step} & \textbf{Name} & \textbf{Description} \\
\hline
0 & Keyword cues & Identify self-harm-relevant keywords or phrases in the note. \\
1 & Act on body & Determine whether a physical act on the body is described. \\
2 & Injury present & Determine whether a physical injury is documented. \\
3 & Method & Identify the self-harm method if reported from a predefined method list. \\
4 & Timing & Assess the timing of the act relative to presentation. \\
5 & Intent & Assess whether the act was intentional or accidental. \\
-- & Summary & Synthesise the preceding steps into a final binary self-harm judgement. \\
\hline
\end{tabularx}
\end{table}

\paragraph{Stage 3: Evidence-Augmented Classifier}
Stage 3 is the final supervised classification stage. It combines the original triage note with the evidence and reasoning extracted in Stage 2 to generate the final prediction. By augmenting the raw note with evidence quotations and free-text reasoning, this stage aims to make relevant self-harm signals more explicit and less dependent on site-specific wording. This augmentation is designed to support more robust classification across hospital settings.

\subsection{Experimental Design}

\subsubsection{Data Splits}
Table \ref{tab:dataset} summarises the dataset characteristics across hospitals. RMH was used as the development site. RMH records from 2012-2017 were split into development and test sets using an 80/20 ratio. The RMH development set was further divided into four chunks stratified by self-harm label to support sequential pipeline development while preventing data leakage between stages. Records with empty triage text were excluded before model input, resulting in small differences in final chunk sizes. The RMH test set was used for the final in-distribution evaluation. RMH records from 2018-2022 were used as a prospective test set.

For external validation, LRH (2012 to 2017) and SunH (2019 to 2022) records were split 20/80 into development and test sets, stratified by self-harm label. A smaller development split was used at the external sites to preserve most records for external evaluation. LRH records from 2018 to 2022 provided an additional prospective external test set.

\subsubsection{Stages 1 and 2: Prompt development}

Stages 1 and 2 used \texttt{gpt-oss-20b}~\cite{agarwal2025gpt}, an open-weight model released by OpenAI, deployed locally on institutional high-performance computing infrastructure using a single NVIDIA A100 80GB GPU. All prompts were developed on RMH Chunks 1 and 2 and fixed before any test-set evaluation. To improve output stability, both Stage 1 and Stage 2 used repeated LLM inference with non-zero temperature. For each note, Stage 1 used three independent runs and Stage 2 used five independent runs. The final label at each stage was determined by majority vote across runs, following the self-consistency prompting approach \cite{wang2023selfconsistencyimproveschainthought}. The full Stage 1 and Stage 2 prompts are provided in Supplementary material Text 1, and detailed inference settings are provided in Supplementary material Table 5.

\subsubsection{Stage 3: Model Development}

\paragraph{Feature Construction}
Stage 3 used features derived from both the original triage note and the Stage 2 outputs, as shown in Table \ref{tab:feature_blocks}. All TF-IDF vectorisers were fitted on RMH Chunk 3 and applied without refitting to the held-out test sets. While AUPRC was used as the primary evaluation metric, feature and model selection also considered F1 at the selected operating threshold, as the final model was applied using a predefined threshold to produce binary labels. 
% When candidate configurations had similar AUPRC, we selected the one with better threshold-level F1.

\begin{table}[ht]
\centering
\small
\setlength{\tabcolsep}{1.5pt}
\caption{Features used in Stage 3.}
\label{tab:feature_blocks}
\begin{tabularx}{\linewidth}{
>{\raggedright\arraybackslash}l
>{\raggedright\arraybackslash}p{2.1cm}
>{\raggedright\arraybackslash}X
>{\raggedright\arraybackslash}X
}
\hline
\textbf{Features} & \textbf{Source} & \textbf{Descriptions} & \textbf{Specification / use} \\
\hline
TF-IDF$_{\text{note}}$ 
& Triage note 
& Lexical patterns in the raw note 
& 5{,}000 unigram features with sublinear TF scaling \\

MiniLM$_{\text{note}}$ 
& Triage note 
& Sentence-level semantic information in the raw note 
& \texttt{all-MiniLM-L6-v2}; 384 dimensions \\

TF-IDF$_{\text{ev}}$ 
& Stage 2 evidence text 
& Explicit evidence spans and key self-harm-related phrases extracted by the LLM 
& 1{,}000 unigram features; used on selected self-harm feature combinations \\

TF-IDF$_{\text{re}}$ 
& Stage 2 reasoning text 
& Explanatory LLM reasoning patterns to self-harm-related interpretation 
& 1{,}000 unigram features; used on selected self-harm feature combinations \\

\hline
\end{tabularx}
\end{table}

\paragraph{Feature Selection}
Feature selection was performed in two ablations, both evaluated on RMH Chunk 4 with the classifier fixed to logistic regression. 
The first ablation evaluated combinations of all feature blocks, with TF-IDF$_\text{ev}$ and TF-IDF$_\text{re}$ constructed using all self-harm features, to identify the best-performing feature set. 
The second ablation took the best feature set from the first ablation, and replaced the aggregated TF-IDF${_\text{ev}}$ and TF-IDF${_\text{re}}$ features with versions constructed from different combinations of self-harm features (Indicators 0-5 and summary, $2^7-1=127$ subsets). This ablation was used to identify the most informative feature combination.
% Full ablation results are reported in Supplementary material [X].

\paragraph{Classifier Selection}

We further compared classification models on RMH Chunk 4 using the selected Stage 3 feature set. Traditional machine learning classifiers implemented in scikit-learn~\cite{pedregosa2011scikit} included logistic regression, multilayer perceptron (MLP), support vector machine (SVM), gradient boosting, extra trees, and random forest. We also evaluated LightGBM~\cite{ke2017lightgbm} and XGBoost~\cite{chen2016xgboost}. For these models, hyperparameters were selected using grid search.

We additionally evaluated BERT-based classifiers, including BERT-base~\cite{devlin2019bert}, SciBERT~\cite{beltagy2019scibert}, ClinicalBERT~\cite{alsentzer2019publicly}, PubMedBERT~\cite{gu2021pubmedbert}, RoBERTa-base~\cite{liu2019roberta}, BiomedRoBERTa \cite{domains}, and GatorTron~\cite{yang2022gatortron}, under both fine-tuned and frozen embedding settings. Fine-tuned models were trained for 3 epochs with a batch size of 64 and a maximum sequence length of 512, given the relatively small training set.

\paragraph{Final Model and Feature Configuration}
The final Stage 3 model used logistic regression, trained on the selected feature configuration: TF-IDF$_{\text{note}}$ + MiniLM$_{\text{note}}$ + TF-IDF$_{\text{ev}}$ + TF-IDF$_{\text{re}}$, with evidence and reasoning features restricted to the \textit{method + summary} combination. The model was trained on RMH Chunk 3. The classification threshold was selected to maximise F1 on a stratified 80/20 hold-out of RMH Chunk 3 and then applied without modification to all held-out test sets. Detailed Stage 3 classifier training and feature settings are provided in Supplementary material Table 6.

\subsection{Evaluation Design}

\subsubsection{Internal, External and Prospective Evaluation}

The final selected model was evaluated on fully held-out test sets. Internal evaluation used the RMH test set from the same period as the RMH development set (2012-2017). Temporal generalisation at the development site was evaluated on the RMH prospective test set from 2018-2022. In the cross-hospital setting, the model trained on RMH Chunk 3 was applied directly to LRH and SunH without site-specific retraining or threshold recalibration. LRH records from 2018-2022 were additionally used as a prospective external test set to assess temporal generalisation under cross-hospital transfer. Since SunH records spanned only from 2019 to 2022, SunH evaluation reflects both cross-hospital and temporal differences from the RMH development data.

\subsubsection{Same-Hospital Upper-Bound Comparison}

To quantify the performance cost of cross-hospital transfer, we evaluated a same-hospital upper-bound condition for LRH and SunH. In this setting, a separate Stage 3 classifier was trained on the local development split of each site and evaluated on the corresponding local test split. In the same-hospital condition, TF-IDF vectorisers and MiniLM encoders were refitted on the local development split.

% depends on whether report the basline
\subsubsection{Baseline Comparisons}
To assess the contribution of Stages 1 and 2, we compared the proposed three-stage approach with three end-to-end note-only baselines trained directly on the full pre-screening RMH Chunk 3 dataset. The first baseline was logistic regression using TF-IDF${_\text{note}}$ and MiniLM${_\text{note}}$ features, corresponding to the final Stage 3 classifier without any LLM-derived features. The second baseline was ClinicalBERT~\cite{alsentzer2019publicly}, fine-tuned directly on the raw triage note using weighted cross-entropy. The first two baselines correspond to the best-performing traditional ML and BERT-based models from the Stage 3 classifier comparison. We additionally included a gradient boosting baseline adapted from the prior ED triage-note self-harm detection approach by Rozova et al.~\cite{rozova_detection_2021}.
All baselines were evaluated on the same test and prospective test sets as the proposed approach.

\subsubsection{Evaluation Metrics and Computational Cost}
Model performance was evaluated using AUPRC, AUROC, F1, precision, and recall. AUPRC was the primary metric due to the strong class imbalance in ED self-harm surveillance. F1 was also used for feature and classifier selection, as it reflects the precision-recall balance at the classification threshold. For end-to-end evaluation, notes excluded at Stage 1 were assigned a predicted probability of zero so that all metrics reflected the cumulative effect of the full approach.
All metrics were estimated with 95\% bootstrapped confidence intervals using 1{,}000 bootstrap iterations stratified by label, and are reported as point estimates $\pm$ half the confidence interval width.

Computational cost was estimated using CodeCarbon~\cite{benoit_courty_2024_11171501}, including wall-clock runtime, energy consumption, and estimated CO$_2$ emissions. Stage 1 and Stage 2 inference were run on a single NVIDIA A100 80GB GPU, while Stage 3 used a CPU-based logistic regression classifier.

\section*{Data Availability}
Due to ethical and governance restrictions for collected hospital data, the datasets cannot be publicly shared. Aggregate results and synthetic examples are provided in the manuscript and supplementary materials.

\section*{Code Availability}
The code used for this study is available in the GitHub repository: \url{https://github.com/LiuliuChen/ED_SH_surveillance}.


\bibliography{sn-bibliography}% common bib file

@article{gillies2018prevalence,
  title={Prevalence and characteristics of self-harm in adolescents: meta-analyses of community-based studies 1990--2015},
  author={Gillies, Donna and Christou, Maria A and Dixon, Andrew C and Featherston, Oliver J and Rapti, Iro and Garcia-Anguita, Alicia and Villasis-Keever, Miguel and Reebye, Pratibha and Christou, Evangelos and Al Kabir, Nagat and others},
  journal={Journal of the American Academy of Child \& Adolescent Psychiatry},
  volume={57},
  number={10},
  pages={733--741},
  year={2018},
  publisher={Elsevier},
  doi={10.1016/j.jaac.2018.06.018}
}

@article{hawton2012self,
  title={Self-harm and suicide in adolescents},
  author={Hawton, Keith and Saunders, Kate EA and O'Connor, Rory C},
  journal={The Lancet},
  volume={379},
  number={9834},
  pages={2373--2382},
  year={2012},
  publisher={Elsevier}
}

@article{zubrick2016self,
  title={Self-harm: prevalence estimates from the second {Australian} child and adolescent survey of mental health and wellbeing},
  author={Zubrick, Stephen R and Hafekost, Jennifer and Johnson, Sarah E and Lawrence, David and Saw, Suzy and Sawyer, Michael and Ainley, John and Buckingham, William J},
  journal={Australian \& New Zealand Journal of Psychiatry},
  volume={50},
  number={9},
  pages={911--921},
  year={2016},
  publisher={Sage Publications Sage UK: London, England}
}

@article{carroll2014hospital,
  title={Hospital presenting self-harm and risk of fatal and non-fatal repetition: systematic review and meta-analysis},
  author={Carroll, Robert and Metcalfe, Chris and Gunnell, David},
  journal={PLOS ONE},
  volume={9},
  number={2},
  pages={e89944},
  year={2014},
  publisher={Public Library of Science San Francisco, USA}
}

@misc{aihw_selfharm_2024,
  author       = {{Australian Institute of Health and Welfare}},
  title        = {Suicide and self-harm among young people},
  year         = {2024},
  howpublished = {Suicide \& Self-Harm Monitoring},
  url          = {https://www.aihw.gov.au/suicide-self-harm-monitoring/population-groups/young-people/suicide-self-harm-young-people},
  note         = {Accessed: 2026-04-23}
}

@techreport{rmh_annual_2025,
  author      = {{Melbourne Health}},
  title       = {{Royal} {Melbourne} {Hospital} Annual Report 2024--2025},
  institution = {The Royal Melbourne Hospital},
  year        = {2025},
  url         = {https://www.thermh.org.au/files/documents/Corporate/rmh-annual-report-2024-2025.pdf}
}

@techreport{lrh_annual_2025,
  author      = {{Latrobe Regional Health}},
  title       = {{Latrobe} {Regional} {Health} Annual Report 2024--2025},
  institution = {Latrobe Regional Health},
  year        = {2025},
  url         = {https://lrh.com.au/wp-content/uploads/2025/11/LRH-Annual-Report_2025-Web-Spreads.pdf}
}

@techreport{wh_annual_2024,
  author      = {{Western Health}},
  title       = {Western Health Annual Report 2023--2024},
  institution = {Western Health},
  year        = {2024},
  url         = {https://www.parliament.vic.gov.au/4962d7/globalassets/tabled-paper-documents/tabled-paper-8807/western-health-annual-report-2023-2024.pdf}
}

@article{kuramoto2017detecting,
  title={Detecting suicide-related emergency department visits among adults using the District of {Columbia} syndromic surveillance system},
  author={Kuramoto-Crawford, S Janet and Spies, Erica L and Davies-Cole, John},
  journal={Public Health Reports},
  volume={132},
  number={1\_suppl},
  pages={88S--94S},
  year={2017},
  publisher={SAGE Publications Sage CA: Los Angeles, CA}
}

@article{silva_characteristics_2024,
    title = {Characteristics of surveillance systems for suicide and self-harm: {A} scoping review},
    volume = {4},
    issn = {2767-3375},
    shorttitle = {Characteristics of surveillance systems for suicide and self-harm},
    url = {https://journals.plos.org/globalpublichealth/article?id=10.1371/journal.pgph.0003292},
    doi = {10.1371/journal.pgph.0003292},
    language = {en},
    number = {7},
    urldate = {2026-04-23},
    journal = {PLOS Global Public Health},
    publisher = {Public Library of Science},
    author = {Silva, Aline Conceição and Vanzela, Amanda Sarah and Pedrollo, Laysa Fernanda Silva and Baker, John and Carvalho, José Carlos Marques de and Sequeira, Carlos Alberto da Cruz and Vedana, Kelly Graziani Giacchero and Santos, José Carlos Pereira dos},
    month = jul,
    year = {2024},
    pages = {e0003292},
}

@article{gill2017emergency,
  title={Emergency department as a first contact for mental health problems in children and youth},
  author={Gill, Peter J and Saunders, Natasha and Gandhi, Sima and Gonzalez, Alejandro and Kurdyak, Paul and Vigod, Simone and Guttmann, Astrid},
  journal={Journal of the American Academy of Child \& Adolescent Psychiatry},
  volume={56},
  number={6},
  pages={475--482},
  year={2017},
  publisher={Elsevier}
}

@article{williams_establishing_2015,
    title = {Establishing a self-harm surveillance register to improve care in a general hospital},
    volume = {4},
    url = {https://www.magonlinelibrary.com/doi/abs/10.12968/bjmh.2015.4.1.20},
    doi = {10.12968/bjmh.2015.4.1.20},
    number = {1},
    urldate = {2026-04-23},
    journal = {British Journal of Mental Health Nursing},
    publisher = {Mark Allen Group},
    author = {Williams, Salena},
    month = jan,
    year = {2015},
    pages = {20--25},
}

@article{sveticic2020suicidal,
  title={Suicidal and self-harm presentations to Emergency Departments: The challenges of identification through diagnostic codes and presenting complaints},
  author={Sveticic, Jerneja and Stapelberg, Nicholas CJ and Turner, Kathryn},
  journal={Health Information Management Journal},
  volume={49},
  number={1},
  pages={38--46},
  year={2020},
  publisher={Sage Publications Sage UK: London, England}
}

@article{randall2017emergency,
  title={Emergency department and inpatient coding for self-harm and suicide attempts: validation using clinician assessment data},
  author={Randall, Jason R and Roos, Leslie L and Lix, Lisa M and Katz, Laurence Y and Bolton, James M},
  journal={International Journal of Methods in Psychiatric Research},
  volume={26},
  number={3},
  pages={e1559},
  year={2017},
  publisher={Wiley Online Library}
}

@article{rozova_detection_2021,
    title = {Detection of self-harm and suicidal ideation in emergency department triage notes},
    volume = {29},
    issn = {1067-5027},
    url = {https://pmc.ncbi.nlm.nih.gov/articles/PMC8800520/},
    doi = {10.1093/jamia/ocab261},
    number = {3},
    urldate = {2026-04-22},
    journal = {Journal of the American Medical Informatics Association},
    author = {Rozova, Vlada and Witt, Katrina and Robinson, Jo and Li, Yan and Verspoor, Karin},
    month = dec,
    year = {2021},
    pages = {472--480},
}

@article{obeid2020identifying,
  title={Identifying and predicting intentional self-harm in {Electronic} {Health} {Record} clinical notes: deep learning approach},
  author={Obeid, Jihad S and Dahne, Jennifer and Christensen, Sean and Howard, Samuel and Crawford, Tami and Frey, Lewis J and Stecker, Tracy and Bunnell, Brian E},
  journal={JMIR Medical Informatics},
  volume={8},
  number={7},
  pages={e17784},
  year={2020},
  publisher={JMIR Publications Toronto, Canada}
}

@article{ayre2021developing,
  title={Developing a natural language processing tool to identify perinatal self-harm in {Electronic} {Healthcare} {Records}},
  author={Ayre, Karyn and Bittar, Andr{\'e} and Kam, Joyce and Verma, Somain and Howard, Louise M and Dutta, Rina},
  journal={PLOS ONE},
  volume={16},
  number={8},
  pages={e0253809},
  year={2021},
  publisher={Public Library of Science San Francisco, CA USA}
}

@misc{rozova_portability_2025,
    title = {Portability of an artificial intelligence model for self-harm detection across hospital settings},
    copyright = {© 2025, Posted by Cold Spring Harbor Laboratory. The copyright holder for this pre-print is the author. All rights reserved. The material may not be redistributed, re-used or adapted without the author's permission.},
    url = {https://www.medrxiv.org/content/10.1101/2025.07.10.25331160v1},
    doi = {10.1101/2025.07.10.25331160},
    language = {en},
    urldate = {2026-04-26},
    publisher = {medRxiv},
    author = {Rozova, Vlada and Witt, Katrina and Conway, Mike and Robinson, Jo and Verspoor, Karin},
    month = jul,
    year = {2025},
    note = {ISSN: 3067-2007
Pages: 2025.07.10.25331160},
}

@article{cusick_portability_2022,
    title = {Portability of natural language processing methods to detect suicidality from clinical text in {US} and {UK} electronic health records},
    volume = {10},
    issn = {2666-9153},
    url = {https://www.sciencedirect.com/science/article/pii/S2666915322001226},
    doi = {10.1016/j.jadr.2022.100430},
    urldate = {2026-04-26},
    journal = {Journal of Affective Disorders Reports},
    author = {Cusick, Marika and Velupillai, Sumithra and Downs, Johnny and Campion, Thomas R. and Sholle, Evan T. and Dutta, Rina and Pathak, Jyotishman},
    month = dec,
    year = {2022},
    pages = {100430},
}

@article{holmes_applications_2025,
    title = {Applications of {Large} {Language} {Models} in the Field of Suicide Prevention: Scoping Review},
    volume = {27},
    shorttitle = {Applications of {Large} {Language} {Models} in the {Field} of {Suicide} {Prevention}},
    url = {https://www.jmir.org/2025/1/e63126},
    doi = {10.2196/63126},
    language = {EN},
    number = {1},
    urldate = {2026-04-26},
    journal = {Journal of Medical Internet Research},
    publisher = {JMIR Publications Inc., Toronto, Canada},
    author = {Holmes, Glenn and Tang, Biya and Gupta, Sunil and Venkatesh, Svetha and Christensen, Helen and Whitton, Alexis},
    month = jan,
    year = {2025},
    pages = {e63126},
}

@article{barak-corren_validation_2020,
    title = {Validation of an {Electronic} {Health} {Record}-Based Suicide Risk Prediction Modeling Approach Across Multiple Health Care Systems},
    volume = {3},
    issn = {2574-3805},
    doi = {10.1001/jamanetworkopen.2020.1262},
    language = {eng},
    number = {3},
    journal = {JAMA Network Open},
    author = {Barak-Corren, Yuval and Castro, Victor M. and Nock, Matthew K. and Mandl, Kenneth D. and Madsen, Emily M. and Seiger, Ashley and Adams, William G. and Applegate, R. Joseph and Bernstam, Elmer V. and Klann, Jeffrey G. and McCarthy, Ellen P. and Murphy, Shawn N. and Natter, Marc and Ostasiewski, Brian and Patibandla, Nandan and Rosenthal, Gary E. and Silva, George S. and Wei, Kun and Weber, Griffin M. and Weiler, Sarah R. and Reis, Ben Y. and Smoller, Jordan W.},
    month = mar,
    year = {2020},
    pages = {e201262},
}

@article{agarwal2025gpt,
  title={gpt-oss-120b \& gpt-oss-20b model card},
  author={Agarwal, Sandhini and Ahmad, Lama and Ai, Jason and Altman, Sam and Applebaum, Andy and Arbus, Edwin and Arora, Rahul K and Bai, Yu and Baker, Bowen and Bao, Haiming and others},
  journal={arXiv preprint arXiv:2508.10925},
  year={2025}
}

@misc{wang2023selfconsistencyimproveschainthought,
      title={Self-Consistency Improves Chain of Thought Reasoning in Language Models}, 
      author={Xuezhi Wang and Jason Wei and Dale Schuurmans and Quoc Le and Ed Chi and Sharan Narang and Aakanksha Chowdhery and Denny Zhou},
      year={2023},
      eprint={2203.11171},
      archivePrefix={arXiv},
      primaryClass={cs.CL},
      url={https://arxiv.org/abs/2203.11171}, 
}

@inproceedings{beltagy2019scibert,
  title={{SciBERT}: A Pretrained Language Model for Scientific Text},
  author={Beltagy, Iz and Lo, Kyle and Cohan, Arman},
  booktitle={Proceedings of the 2019 Conference on Empirical Methods in Natural Language Processing and the 9th International Joint Conference on Natural Language Processing (EMNLP-IJCNLP)},
  pages={3615--3620},
  year={2019},
  publisher={Association for Computational Linguistics},
  doi={10.18653/v1/D19-1371}
}

@inproceedings{alsentzer2019publicly,
    title = "Publicly Available Clinical {BERT} Embeddings",
    author = "Alsentzer, Emily  and
      Murphy, John  and
      Boag, William  and
      Weng, Wei-Hung  and
      Jindi, Di  and
      Naumann, Tristan  and
      McDermott, Matthew",
    booktitle = "Proceedings of the 2nd Clinical Natural Language Processing Workshop",
    month = jun,
    year = "2019",
    address = "Minneapolis, Minnesota, USA",
    publisher = "Association for Computational Linguistics",
    url = "https://aclanthology.org/W19-1909/",
    doi = "10.18653/v1/W19-1909",
    pages = "72--78",
}

@inproceedings{devlin2019bert,
  title={{BERT}: Pre-training of Deep Bidirectional Transformers for Language Understanding},
  author={Devlin, Jacob and Chang, Ming-Wei and Lee, Kenton and Toutanova, Kristina},
  booktitle={Proceedings of the 2019 Conference of the North American Chapter of the Association for Computational Linguistics: Human Language Technologies, Volume 1 (Long and Short Papers)},
  pages={4171--4186},
  year={2019},
  address={Minneapolis, Minnesota, USA},
  publisher={Association for Computational Linguistics},
  doi={10.18653/v1/N19-1423}
}

@article{gu2021pubmedbert,
  title={Domain-Specific Language Model Pretraining for Biomedical Natural Language Processing},
  author={Gu, Yu and Tinn, Robert and Cheng, Hao and Lucas, Michael and Usuyama, Naoto and Liu, Xiaodong and Naumann, Tristan and Gao, Jianfeng and Poon, Hoifung},
  journal={ACM Transactions on Computing for Healthcare},
  volume={3},
  number={1},
  pages={1--23},
  year={2021},
  doi={10.1145/3458754}
}

@article{liu2019roberta,
  title={{RoBERTa}: A Robustly Optimized {BERT} Pretraining Approach},
  author={Liu, Yinhan and Ott, Myle and Goyal, Naman and Du, Jingfei and Joshi, Mandar and Chen, Danqi and Levy, Omer and Lewis, Mike and Zettlemoyer, Luke and Stoyanov, Veselin},
  journal={arXiv preprint arXiv:1907.11692},
  year={2019},
  doi={10.48550/arXiv.1907.11692}
}

@article{yang2022gatortron,
  title={A {Large} {Language} {Model} for {Electronic} {Health} {Records}},
  author={Yang, Xi and Chen, Aibing and PourNejatian, Nima and Shin, Hoo-Chang and Smith, Kaleb E. and Parisien, Christopher and Compas, Chris and Martin, Cheryl and Flores, Michael and Zhang, Ying and Magoc, Tomas and Harle, Christopher A. and Lipori, George and Mitchell, Demetrius and Hogan, William R. and Shenkman, Elizabeth and Bian, Jiang},
  journal={npj Digital Medicine},
  volume={5},
  number={1},
  pages={194},
  year={2022},
  doi={10.1038/s41746-022-00742-2}
}

@inproceedings{ke2017lightgbm,
author = {Ke, Guolin and Meng, Qi and Finley, Thomas and Wang, Taifeng and Chen, Wei and Ma, Weidong and Ye, Qiwei and Liu, Tie-Yan},
title = {{LightGBM}: a highly efficient gradient boosting decision tree},
year = {2017},
isbn = {9781510860964},
booktitle = {Proceedings of the 31st International Conference on Neural Information Processing Systems},
pages = {3149--3157},
numpages = {9},
location = {Long Beach, California, USA},
series = {NIPS'17},
publisher = {Curran Associates, Inc.}
}

@inproceedings{chen2016xgboost,
author = {Chen, Tianqi and Guestrin, Carlos},
title = {{XGBoost}: A Scalable Tree Boosting System},
year = {2016},
isbn = {9781450342322},
publisher = {Association for Computing Machinery},
address = {New York, NY, USA},
url = {https://doi.org/10.1145/2939672.2939785},
doi = {10.1145/2939672.2939785},
booktitle = {Proceedings of the 22nd ACM SIGKDD International Conference on Knowledge Discovery and Data Mining},
pages = {785--794},
numpages = {10},
location = {San Francisco, California, USA},
series = {KDD '16}
}

@article{pedregosa2011scikit,
  title={Scikit-learn: Machine Learning in {Python}},
  author={Pedregosa, Fabian and Varoquaux, Ga\"el and Gramfort, Alexandre and Michel, Vincent and Thirion, Bertrand and Grisel, Olivier and Blondel, Mathieu and Prettenhofer, Peter and Weiss, Ron and Dubourg, Vincent and Vanderplas, Jake and Passos, Alexandre and Cournapeau, David and Brucher, Matthieu and Perrot, Matthieu and Duchesnay, \'Edouard},
  journal={Journal of Machine Learning Research},
  volume={12},
  pages={2825--2830},
  year={2011}
}

@inproceedings{domains,
    title = "Don{'}t Stop Pretraining: Adapt Language Models to Domains and Tasks",
    author = "Gururangan, Suchin  and
      Marasovi{\'c}, Ana  and
      Swayamdipta, Swabha  and
      Lo, Kyle  and
      Beltagy, Iz  and
      Downey, Doug  and
      Smith, Noah A.",
    booktitle = "Proceedings of the 58th Annual Meeting of the Association for Computational Linguistics",
    month = jul,
    year = "2020",
    address = "Online",
    publisher = "Association for Computational Linguistics",
    url = "https://aclanthology.org/2020.acl-main.740/",
    doi = "10.18653/v1/2020.acl-main.740",
    pages = "8342--8360"
}

@article{ross2023emergency,
  title={Emergency department presentations with suicide and self-harm ideation: a missed opportunity for intervention?},
  author={Ross, E and Murphy, S and O’Hagan, D and Maguire, A and O’Reilly, D},
  journal={Epidemiology and Psychiatric Sciences},
  volume={32},
  pages={e24},
  year={2023}
}

@article{Malleyicd,
author = {O'Malley, Kimberly J. and Cook, Karon F. and Price, Matt D. and Wildes, Kimberly Raiford and Hurdle, John F. and Ashton, Carol M.},
title = {Measuring Diagnoses: {ICD} Code Accuracy},
journal = {Health Services Research},
volume = {40},
number = {5p2},
pages = {1620-1639},
doi = {https://doi.org/10.1111/j.1475-6773.2005.00444.x},
url = {https://onlinelibrary.wiley.com/doi/abs/10.1111/j.1475-6773.2005.00444.x},
eprint = {https://onlinelibrary.wiley.com/doi/pdf/10.1111/j.1475-6773.2005.00444.x},
year = {2005}
}

@misc{benoit_courty_2024_11171501,
  author       = {Benoit Courty and
                  Victor Schmidt and
                  Sasha Luccioni and
                  Goyal-Kamal and
                  MarionCoutarel and
                  Boris Feld and
                  Jérémy Lecourt and
                  LiamConnell and
                  Amine Saboni and
                  Inimaz and
                  supatomic and
                  Mathilde Léval and
                  Luis Blanche and
                  Alexis Cruveiller and
                  ouminasara and
                  Franklin Zhao and
                  Aditya Joshi and
                  Alexis Bogroff and
                  Hugues de Lavoreille and
                  Niko Laskaris and
                  Edoardo Abati and
                  Douglas Blank and
                  Ziyao Wang and
                  Armin Catovic and
                  Marc Alencon and
                  Michał Stęchły and
                  Christian Bauer and
                  Lucas Otávio N. de Araújo and
                  JPW and
                  MinervaBooks},
  title        = {mlco2/codecarbon: v2.4.1},
  month        = may,
  year         = 2024,
  publisher    = {Zenodo},
  version      = {v2.4.1},
  doi          = {10.5281/zenodo.11171501},
  url          = {https://doi.org/10.5281/zenodo.11171501}
}

@article{bandara2022surveillance,
author = {Bandara, Piumee and Page, Andrew and Hammond, Trent Ernest and Sperandei, Sandro and Stevens, Garry John and Gunja, Naren and Anand, Manish and Jones, Alison and Carter, Greg},
title = {Surveillance of Hospital-Presenting Intentional Self-Harm in Western Sydney, Australia, During the Implementation of a New Self-Harm Reporting Field},
journal = {Crisis},
volume = {44},
number = {2},
pages = {135-145},
year = {2023},
doi = {10.1027/0227-5910/a000845},
    note ={PMID: 35138153},

URL = { 
        https://doi.org/10.1027/0227-5910/a000845
},
}

@article{hunter,
author = {Whyte, Ian M and Dawson, Andrew H and Carter, Gregory L and Levey, Catherine M and Buckley, Nicholas A},
title = {A model for the management of self-poisoning},
journal = {Medical Journal of Australia},
volume = {167},
number = {3},
pages = {142-146},
doi = {https://doi.org/10.5694/j.1326-5377.1997.tb138813.x},
url = {https://onlinelibrary.wiley.com/doi/abs/10.5694/j.1326-5377.1997.tb138813.x},
eprint = {https://onlinelibrary.wiley.com/doi/pdf/10.5694/j.1326-5377.1997.tb138813.x},
year = {1997}
}

\section*{Funding}
KW is supported by a Dame Kate Campbell Fellowship from the Faculty of Medicine, Dentistry, and Health Sciences at The University of Melbourne and the Victorian Collaborative Centre for Mental Health and Wellbeing. JR is supported by an NHMRC Investigator Grant (2008460) and the University of Melbourne Dame Kate Campbell Fellowship.

\section*{Acknowledgements}
We thank Associate Professor Jonathan Knott and Owen Conolly for facilitating data acquisition at the Royal Melbourne Hospital and the Latrobe Regional Health. We also thank Hannah Richards and Lu Zhang for their assistance with manual data coding. 

\section*{Author Contributions}
LC, MC, and VR contributed to the conception and design of the study. GR, EB, KW, ML, and JR contributed to data acquisition, annotation, and/or interpretation of the data. ML, KW, and JR managed stakeholder relationships and contributed to funding acquisition. GR, EB, KW, ML, and JR contributed domain expertise in self-harm and suicide prevention. LC performed the data analysis and implemented the approach used in the study, with supervision and input from MC and VR. LC drafted the manuscript. GR, EB, KW, ML, JR, MC, and VR reviewed and revised the manuscript. All authors approved the submitted version and agreed to be accountable for their own contributions and for ensuring that questions related to the accuracy or integrity of any part of the work are appropriately investigated and resolved.

\section*{Competing Interests}
The authors declare no competing interests.

% \begin{appendices}

% \section{Appendix A}\label{secA1}

% An appendix contains Supplementary material information that is not an essential part of the text itself but which may be helpful in providing a more comprehensive understanding of the research problem or it is information that is too cumbersome to be included in the body of the paper.

% \end{appendices}

\end{document}